\pdfoutput=1

\documentclass[11pt]{article}

\usepackage{acl}

\usepackage{times}
\usepackage{latexsym}

\usepackage[T1]{fontenc}

\usepackage[utf8]{inputenc}

\usepackage{microtype}

\usepackage{algorithmic}
\usepackage{algorithm}

\usepackage{amsmath,amsfonts,bm}

\def\eqref#1{equation~\ref{#1}}

\def\1{\bm{1}}

\DeclareMathAlphabet{\mathsfit}{\encodingdefault}{\sfdefault}{m}{sl}
\SetMathAlphabet{\mathsfit}{bold}{\encodingdefault}{\sfdefault}{bx}{n}

\usepackage{wrapfig}
\usepackage{hyperref}
\usepackage{url}
\usepackage{caption}
\usepackage{subcaption}
\usepackage{graphicx}
\usepackage{amsmath,amssymb,amsfonts,amsbsy,amsfonts,latexsym}
\usepackage{booktabs}
\usepackage{adjustbox}
\usepackage{enumitem}
\usepackage{xparse}
\usepackage{multirow}
\usepackage{array, caption, makecell}
\usepackage{xspace}
\usepackage{paralist}
\usepackage{booktabs}

\usepackage{soul}

\usepackage{textcomp}

\newcommand\appref{Appendix~\ref}
\newcommand\eref{Eq.~\ref}
\newcommand\fref{Figure~\ref}
\newcommand\tref{Table~\ref}
\newcommand\sref{Section~\ref}

\newcommand{\new}[1]{#1}
\newcommand{\orimodel}{\phi\xspace}
\newcommand{\growth}{\mathbb{G}\xspace}
\newcommand{\gptlarge}{GPT2$_{\textsc{Large}}$\xspace}
\newcommand{\gptbase}{GPT2$_{\textsc{Base}}$\xspace}

\newcommand{\pregrowth}{\textsc{pre-growth}\xspace}
\newcommand{\postgrowth}{\textsc{post-growth}\xspace}
\newcommand{\optimality}{\textsc{optimality}\xspace}
\newcommand{\growthtarget}{\textsc{growth-target}\xspace}
\newcommand{\model}{\phi}

\newcommand{\lossgrowth}{L_{\mathbb{G}}}

\newcommand{\trainingdynamics}{\frac{\partial \mathcal{L}}{\partial C}}

\newcommand{\optimalthreshold}{\tau_{opt}}
\newcommand{\growththreshold}{\tau_{\mathbb{G}}}

\title{Staged Training for Transformer Language Models}

\author{Sheng Shen$^\dagger$ ~~ Pete Walsh$^\diamond$ ~~ Kurt Keutzer$^\dagger$  ~~ Jesse Dodge$^\diamond$ ~~ Matthew Peters$^\diamond$ ~~ Iz Beltagy$^\diamond$ \\
  $^\dagger$ University of California, Berkeley \\
  $^\diamond$ Allen Institute for AI \\
  \texttt{sheng.s@berkeley.edu ~~~ keutzer@eecs.berkeley.edu}\\ \texttt{petew,jessed,matthewp,beltagy@allenai.org}

}

\begin{document}
\maketitle
\let\thefootnote\relax\footnotetext{Corresponding author: beltagy@allenai.org}
\begin{abstract}
The current standard approach to scaling transformer language models trains each model size from a different random initialization.
As an alternative, we consider a staged training setup that begins with a small model and incrementally increases the amount of compute used for training by applying a ``growth operator'' to increase the model depth and width. 
By initializing each stage with the output of the previous one, the training process effectively re-uses the compute from prior stages and becomes more efficient.
Our growth operators each take as input the entire training state (including model parameters, optimizer state, learning rate schedule, etc.) and output a new training state from which training continues.
We identify two important properties of these growth operators, namely that they preserve both the loss and the ``training dynamics'' after applying the operator.
While the loss-preserving property has been discussed previously, to the best of our knowledge this work is the first to identify the importance of preserving the training dynamics (the rate of decrease of the loss during training).
To find the optimal schedule for stages, 
we use the scaling laws from~\cite{kaplan2020scaling} to find a precise schedule that 
gives the most compute saving by starting a new stage when training efficiency starts decreasing.
We empirically validate our growth operators and staged training for autoregressive language models, showing up to 22\%  compute savings compared to a strong baseline trained from scratch.
Our code is available at~\url{https://github.com/allenai/staged-training}.
\end{abstract}


\section{Introduction}

Language models form the backbone of many modern NLP systems, and these language models have become progressively larger in recent years.
Parameter counts for these models have grown significantly from ELMo (94 M) \citep{Peters:2018} to GPT-3 (175 B) \citep{brown2020language}.
While larger models with more learnable parameters perform better on a wide range of tasks, the computational cost to train or even just to evaluate these models has become prohibitively expensive \citep{Schwartz:2020}.
In this paper, we demonstrate a method to reduce the compute cost of training transformer language models \cite{Vaswani2017AttentionIA} through a staged training setup that iteratively builds a large model from a smaller one.

\begin{figure}
\centering
\includegraphics[width=0.35\textwidth]
{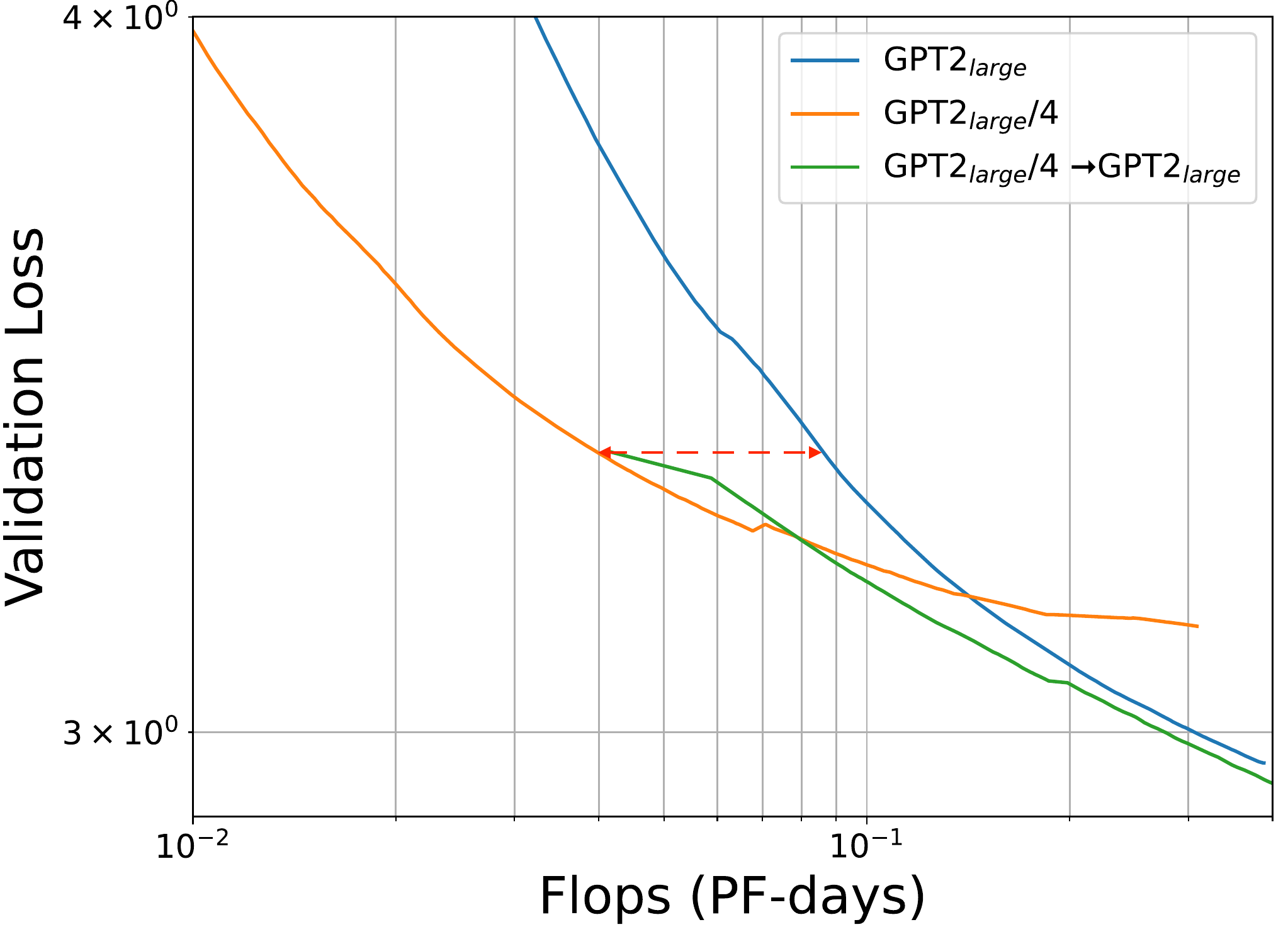}
    \caption{
    We train a GPT2$_\textsc{large}$ (768M parameters) transformer language model by first training a model 1/4 the size (orange line), then increasing the model size by 4x by applying a growth operator to the entire training state, and restarting training (green line).  The result is a large size model with comparable loss to one trained from scratch (blue line) but with reduced compute cost illustrated initially by the dashed red arrow.
    }
    \label{fig:optimal_schedule}
\end{figure}


Most prior work on scaling language models initializes each model size from a random initialization and trains to convergence \cite{kaplan2020scaling}.
This work illustrated an intriguing property of model training shown in \fref{fig:optimal_schedule}, namely that smaller models are initially more compute efficient then larger models, but eventually the larger model will reach a lower loss.
Our central idea is to take advantage of this property by first training a smaller model in the compute efficient region, applying a growth operator to the entire training state, and restarting training with a larger sized model.
We introduce two operators to perform this growing operation along the model depth and width dimensions.
We also identify two important properties of the operators: first, the loss before growing the model is preserved, and second, the training dynamics of the grown model match that of an equivalent model trained from scratch.
To maintain training dynamics, growth operators must take into the entire training state, including the optimizer state and learning rate schedule, in addition to the model weights.
As can be seen in \fref{fig:optimal_schedule} our growth operator is loss-preserving, and we show in subsequent sections that it also preserves the training dynamics.
These properties also make applying the growth operators conceptually and algorithmically simple as they won't disrupt the training process, and prior results regarding the model size needed to converge to a particular loss still hold. 


While prior work~\cite{rusu2016progressive,wei2016network,gong2019efficient,liu2019splitting,li2020train,press2020shortformer,gu2021transformer,li2021curriculum,rae2021scaling,evci2022gradmax} has examined some aspects of staged training, our work is the first to address all aspects including how to grow the entire training state and set the stage schedule.
We begin by describing the details of our growth operators for the model weights and optimizer state.
We then present a principled way to chose the stage schedule, including how to chose the model sizes and number of gradient for each stage.
Intuitively, we should start a new stage when the training efficiency decreases and the rate of loss decrease starts to slow down.
To formalize this intuition, we use the scaling laws
 from~\cite{kaplan2020scaling} to find the optimal schedule that gives the maximum compute saving.
 We then show how to approximate the optimal schedule in the realistic scenario without perfect knowledge of the scaling laws.
 We empirically validate our approach with GPT2 style \cite{radford2019language} auto-regressive language model models and demonstrate 5-30 \% compute savings measured by validation loss, and zero-shot perplexity using two benchmark datasets.



\begin{figure*}[t]
    \centering
    \begin{subfigure}{.35\textwidth}
        \centering
        \includegraphics[width=\textwidth]{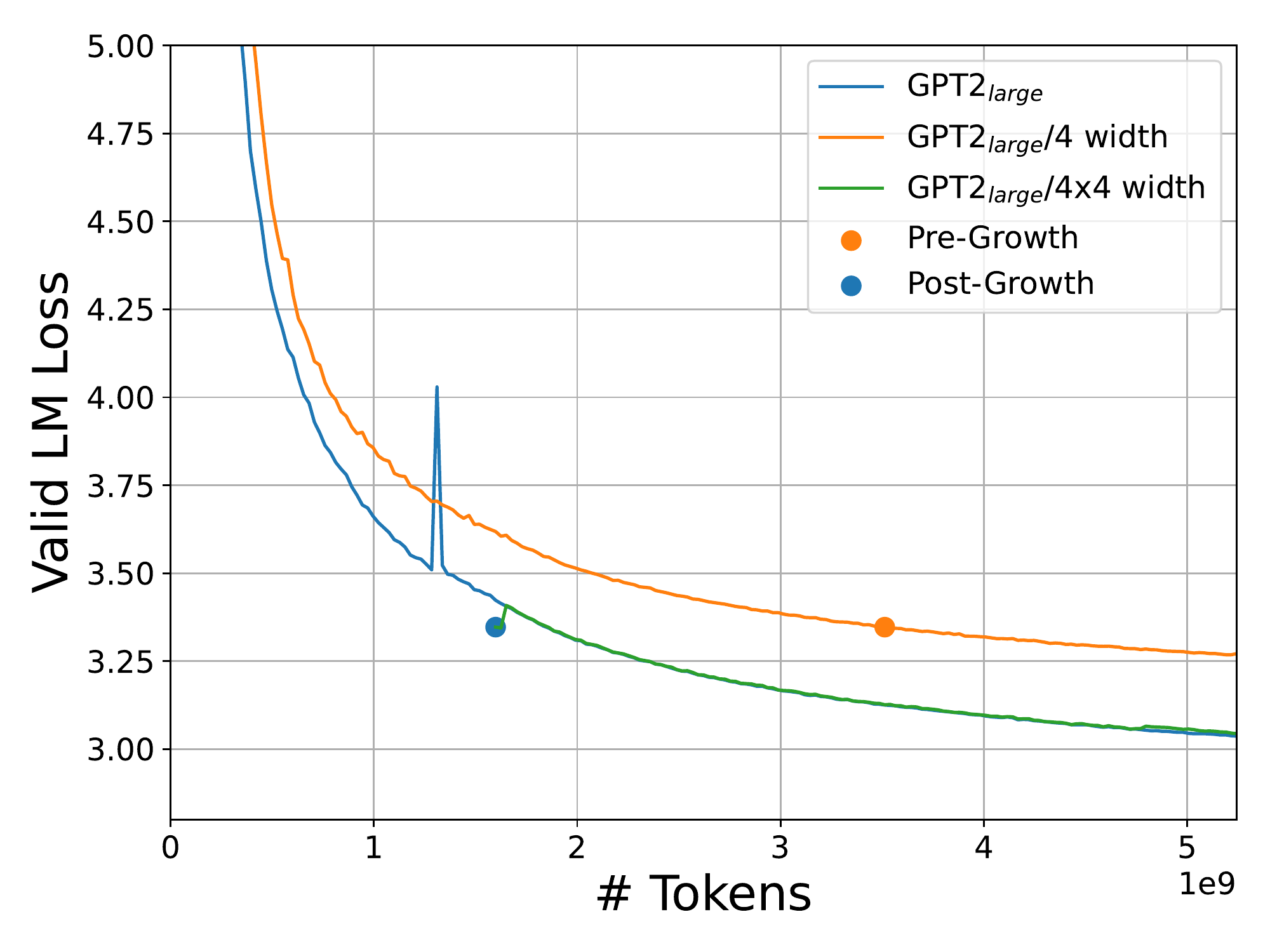}
        \caption{Width growth with GPT2$_\textsc{large}$ as target}
        \label{fig:width_depth_token_a}
    \end{subfigure}
    \begin{subfigure}{.35\textwidth}
        \centering
        \includegraphics[width=\textwidth]{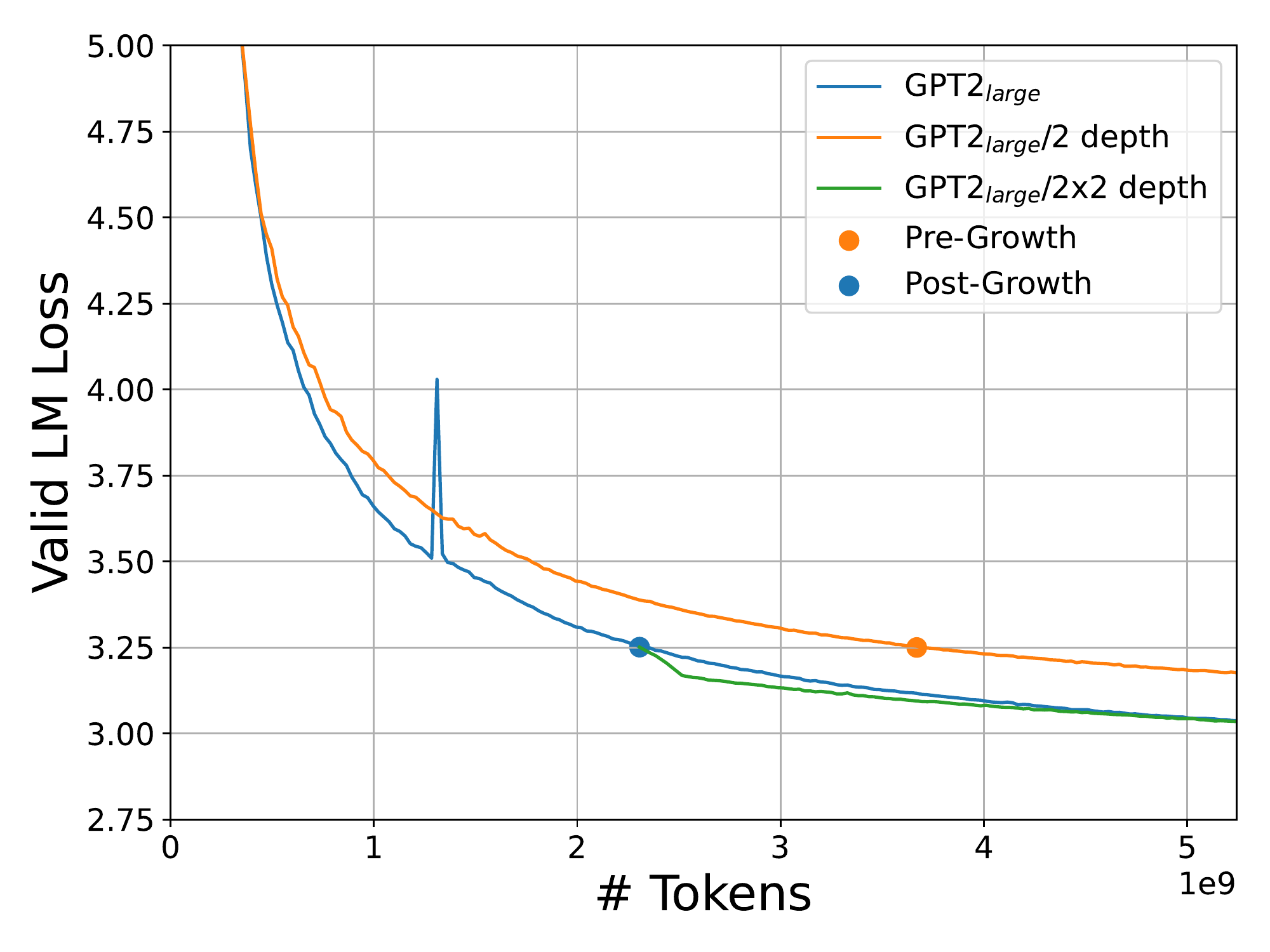}
        \caption{Depth growth with GPT2$_\textsc{large}$ as target}
    \end{subfigure}
    \caption{
    Our growth operators are loss-preserving and training-dynamics preserving.
    Using (a) as an example, {GPT2$_\text{\textsc{large}/4}$} is the original model which is 4x smaller than the target model {GPT2$_\textsc{large}$}. The model {GPT2$_\text{\textsc{large}/4x4}$} is the grown model resulting from growing {GPT2$_\text{\textsc{large}/4}$} by 4x (doubling model width). The \pregrowth point is highlighted on the original model {GPT2$_\text{\textsc{large}/4}$}, 
    and the \postgrowth point is highlighted on the grown model  {GPT2$_\text{\textsc{large}/4x4}$}. 
    The \pregrowth and \postgrowth points have the same loss, showing that the width growth operator is loss-preserving. 
    To demonstrate that it is also training dynamics-preserving,
    we overlay the loss curve for the grown model over the target model and confirm the rate of loss decrease with respect to the number of tokens is the same as the target model trained from scratch.
    The x-axis is number of training tokens since random initialization, or from the start of the training stage (for {GPT2$_\text{\textsc{large}/4x4}$}).
    A similar result is seen in (b) for the depth growth operator.
    }
    \label{fig:width_depth_token}
\end{figure*}

\section{Definitions and Properties of Growth Operators}
\label{sec:properties}

\subsection{Definitions}
We begin by defining some terms which will be used throughout the paper.
We consider a model $\mathbf{y} = \model(\mathbf{x}, \theta)$ which takes input $\mathbf{x}$, outputs $\mathbf{y}$, with parameters $\theta$.
The model is trained by minimizing a loss function $\textit{loss}(\mathbf{y}, \hat{\mathbf{y}}) \in \rm \mathbb{R}$ through a sequence of parameter updates, obtained by running an optimizer.
\new{We will also write $\textit{loss}(\model, \mathcal{D})$ for the total loss over a dataset $\mathcal{D} = \{\mathbf{x}_i, \mathbf{y}_i\}$ (or just $\textit{loss}(\model)$).}
The parameter updates for a particular mini-batch are determined by both the mini-batch and the training state, $\mathcal{T} = \{\theta, (m, v), \lambda(t)\}$, including the model parameters $\theta$, optimizer state ($m, v$, here the first and second moments of the Adam optimizer), and learning rate schedule $\lambda(t)$.
Given a training state, we apply a growth operator $\mathbb{G}(\mathcal{T}_\text{orig}) = \mathcal{T}_\text{grow}$ that takes the original training state and outputs a grown training state where the model size has increased, along with a corresponding compatible optimizer state and learning rate schedule.

\subsection{Desired properties}
In this section, we define two key properties of growth operators.
Building on~\citep{chen2015net2net} we revisit the \textit{loss-preserving} property, and we 
introduce a more challenging property, the \textit{training-dynamic-preserving} property.

\subsubsection{Preserving Loss}
A function-preserving growth operator is one that takes as input an original model and returns a grown model that represents the same function as the original model.  If an operator is function-preserving then it is also loss-preserving.

Mathematically, we can formulate loss-preserving as  
\begin{equation}
\label{eq:loss_preserving}
   \textit{loss}(\mathbb{G}(\model)(\mathbf{x}), \mathbf{y}) = \textit{loss}(\model(\mathbf{x}), \mathbf{y})
\end{equation}
for any $(\mathbf{x}, \mathbf{y})$.
A growth operator that is not loss-preserving wastes time and compute initially until it recovers the same performance of the original model. \fref{fig:width_depth_token} (and the more detailed \fref{fig:width_depth_token_base}) show examples of 
the proposed width and depth growth operators being loss-preserving. 

\subsubsection{Preserving Training Dynamics}

We define the training dynamics as the rate of decrease of loss relative to the amount of compute, and a training-dynamics-preserving growth operator is one that allows the grown model's loss curve to match that of a target model (a model of the same size as the grown model but trained from scratch).

\new{
Formally, let $\model_{k+1}$ be the resulting model after applying the optimizer update to model $\model_k$ with training state $\mathcal{T}_k$.  Applying the update requires some amount of compute $C_k$.  The training process produces a loss curve that associates the loss with the total amount of compute used for training
\[
\mathcal{L}(\model, C) = \{(\overline{C}_k, \mathit{loss}(\model_k, \mathcal{D}_k)), \hspace{1ex} k=1, 2, \ldots\}
\]
where $\overline{C}_k = \sum_{i<=k} C_i$ is the total compute used at step $k$.
The training dynamics is the compute efficiency of training, the expected decrease in the loss relative to the amount of compute:
\[
\frac{\partial }{\partial C}\mathcal{L}(\model, C) = \mathbb{E}_{D}[ \frac{\mathit{loss}(\model_{k}, \mathcal{D}) - \mathit{loss}(\model_{k+1}, \mathcal{D})}{C_k} ]\]
which we denote by $\frac{\partial \mathcal{L}}{\partial C}$ as an abuse of notation.
Practically it is easy to estimate during training by monitoring the model's loss.

We can now define a training-dynamics-preserving growth operator $\mathbb{G}$ at a point on the loss curve with loss $L_\mathbb{G}$ as one that preserves the efficiency of training of the grown model vs.\ a target model trained from scratch:
\begin{equation}
\label{eq:training_dynamics_preserving}
\frac{\partial }{\partial C} \mathcal{L}(\mathbb{G}(\model_{orig}), C) = \frac{\partial }{\partial C} \mathcal{L}(\model_{target}, C)
\end{equation}
where efficiency is evaluated at $L_\mathbb{G}$.


Notice that while loss-preserving is a property comparing the original and grown models, 
training-dynamics-preserving is a property comparing the grown and target models.}
This property makes it possible for the grown model to ``jump'' from the loss curve of the smaller model to the large model and always benefit from faster convergence. 
A growth operator that does not satisfy this requirement could be creating a larger model but with limited capacity or one that is more difficult to train. 
Figure~\ref{fig:width_depth_token} (and the more detailed \fref{fig:width_depth_token_base}) shows examples of the width and depth growth operators preserving the training dynamics, where the grown model perfectly follows the loss curves of the target model.

While the function-preserving property of a growth operator can be confirmed based on the implementation itself, preserving the training dynamics goes beyond just growing the model size; it must be empirically evaluated. To preserve training dynamics, one must address the whole training state including the learning rate and the optimizer state, which are hardly discussed in prior work. We discuss this further in the next section.

\section{Growth Operators}
\label{sec:growth_operator}
We introduce two growth operators below; the operators are described generally, though of course the implementations are model-specific. Our experiments are using the GPT2 transformer  architecture~\cite{radford2019language}. 

\subsection{Width}
\label{sec:width}
Our width operator doubles the hidden dimension of the entire model, and therefore increases the number of parameters by approximately 4x.
This operator applies to all weights in the network including embeddings, feed forward layers, bias, and normalization layers, not just the feed forward layers as in prior work~\citep{gu2021transformer}.
It grows each layer in slightly different ways.
Layer-normalization layers and bias terms with weights $W \in \mathbb{R}^d$ are duplicated:
\begin{equation}
   \notag
    \growth(W) = [W, W].
\end{equation}
where $[\cdot,\cdot]$ represents concatenation and we have overloaded $\growth(\cdot)$ to apply to a weight matrix instead of the entire training state.  Embedding layers are handled in a similar manner.
For the feed forward weights $W \in R^{n x m}$, we design the growth operators as 
\begin{equation*}
\growth(W) =
\begin{pmatrix}
W & Z \\
Z & W \\
\end{pmatrix}
\end{equation*}
where $Z$ is a zero matrix of size similar to $W$.

In this case, the input before the grown last feed forward layer that produces the logits is two times wider and the final logits are two times larger.
To keep the whole network loss-preserving, we divide the grown weights of the last feed forward layer by a factor of two.


\subsection{Depth}
Our depth operator doubles the number of layers, and therefore increases the number of non-embedding parameters by 2x.
Given a model with layers $(\phi_0, \phi_1, \ldots)$, the depth operator adds an identity layer $\model_\text{id}$ after each layer in the original model, so that the grown model has layers $(\phi_0, \model_\text{id}, \phi_1, \model_\text{id}, \ldots)$.
The identity layer is a layer where the input matches the output $\model_\text{id}(\mathbf{x}) = \mathbf{x}$.

To construct the identity layer, we start with the formulation of each layer in GPT2 as two sublayers:
\begin{equation}
    \begin{aligned}
    {\mathbf{x}'} &= {\mathbf{x}}+\text{Attention}(\text{LN}({\mathbf{x}})),  \\
    {\mathbf{y}} &= {\mathbf{x}'}+\text{FFN}(\text{LN}({\mathbf{x}'})) 
    \end{aligned}
        \label{eq:transformer_block}
\end{equation}
where $\bf x \in \rm R^d$ is the input, $\bf y \in \rm R^d$ is the output. 
LN, Attention, and FFN stands for the layer normalization, multi-head attention and feed-forward operations.
We initialize both the scale and bias parameters of each LN in the identity layers to zero, so that $\text{LN}(\mathbf{x}) = \mathbf{0}$.
We also set the bias parameters of all linear layers to zero, which combined with the LN initialization gives $\text{Attention}(\text{LN}({\bf x}))= \mathbf{0}$ and $\text{FFN}(\text{LN}({\bf x'})) =\mathbf{0}$, and the entire layer reduces to an identity layer at initialization.
Overall, the resulting depth growth operator is loss-preserving.

\subsection{Growth operator's impact on training state}
\label{sec:training_state}
In practice, to build a growth operator that preserves training dynamics, we find it important that optimizer state should be grown in a similar way to the model parameters; initial experiments indicated that it can take many training steps to re-estimate the optimizer state, and the initial phase after growth can be unstable. 
\new{
This is expected because training dynamics is the rate of loss change
$\frac{\partial \mathcal{L}}{\partial C}$.
To match the rate of loss change of the target model, the growth operator needs 
to reproduce the same scale of model updates, 
which are controlled by the update rule of the optimizer.}
Using the \textsc{Adam}~\cite{kingma2014adam} optimizer as an example, 
the update rules are
\begin{equation}
    \label{eq:adam_mmt}
    \begin{aligned}
    m_t & = \beta_1*m_{t-1} + (1-\beta_1)g_t \\
    v_t & = \beta_2*v_{t-1} + (1-\beta_2)g_t^2\\
    \theta_{t+1} & = \theta_{t} - \frac{\lambda(t)}{\sqrt{v_t} + \epsilon}m_t
    \end{aligned}
\end{equation}
where  $g$ and $g^2$  are the first-order gradients and the element-wise squared first-order gradients, $m$ is the first moment (average of $g$), $v$ is the second moment (average of $g^2$), $\beta_1, \beta_2, \in [0, 1)$ are the exponential decay rates for the moment estimates, \new{$\epsilon$ is a small constant, }and $t$ is the time-step. 
\new{
For the grown model to be updated at a rate similar to that of the target model,
it needs to match its learning rate $\lambda(t)$ which we discuss in the next section. It also needs to produce an optimizer state $m, v$ that's compatible 
with the gradients of the grown model, $g(\growth(\orimodel))$ and $g^2(\growth(\orimodel))$. 
Given that $m, v$ are averages of $g, g^2$, we 
argue that $m, v$ should be grown with 
growth operators $\growth_m$, $\growth_v$  
that satisfy the following properties:}\footnote{Abusing the notation; \eref{eq:first_mmt}, \ref{eq:second_mmt} are using $g$, $g^2$ as functions to compute gradients of a model, not the gradients themselves.}
\begin{equation}
\label{eq:first_mmt}
    g(\growth_m(\orimodel)) = {\growth}_{m}(g({\orimodel}))
\end{equation}
\begin{equation}
\label{eq:second_mmt}
    {{g^2}(\growth_{v}(\orimodel))} = {\growth_{v}(g^2(\orimodel))}
\end{equation}
The first condition states that the ``gradients of the grown model'' should match ``growing the gradients of the original model''. The second condition is  similar but for the squared gradients.
To satisfy ~\eref{eq:first_mmt} and \ref{eq:second_mmt}, 
the implementations of $\growth_{m}$, $\growth_{v}$ are slightly different from $\growth$. 
For the width growth operator, some of the weights need to be scaled by 0.5x or 0.25x to account for the 2x scaling in the forward pass (see \sref{sec:width}). 
For the depth growth operator, we copy $m$ and $v$ for the original model layers and set $m$ and $v$ to zero for the identity layers.

Along with the loss-preserving property, the training dynamics preserving property ensures that the new optimizer state is compatible with the grown model weights.

\paragraph{Learning rate}
To match training dynamics, the learning rate schedule of the grown model must match that of the target model.
The intuition is that our growth operators allow the model state to ``jump'' from the loss curve of \new{the original model $\mathcal{L}(\model_{orig}, C)$
to $\mathcal{L}(\model_{target}, C)$ at a point with loss $\lossgrowth$. 
Because our growth operator is loss preserving, the loss  $\lossgrowth$ 
defines two points on the loss curves, \pregrowth on 
$\mathcal{L}(\model_{orig}, C)$ and \growthtarget on $\mathcal{L}(\model_{target}, C)$. 
To match the training dynamics of $\mathcal{L}(\model_{target}, C)$,  we}
start training the grown model with a learning rate schedule that matches the target model but starts from \growthtarget. 
We discuss finding that matched \growthtarget point in \sref{sec:optimal_schedule} and~\ref{sec:practical_schedule}.


\section{Optimal Schedule}
\label{sec:optimal_schedule}
Prior work \cite{gu2021transformer, gong2019efficient} used heuristics to determine the training schedule.
In contrast, our goal is to find the optimal training schedule. An optimal training schedule is one that, given a target model size, 
specifies the optimal sequence of 
growth operators, intermediate model sizes, and number of training steps in each stage leading to the most compute saving.
This section will explain our intuition behind our optimal schedule, then 
explains how to mathematically find it. 

\paragraph{Training to \optimality}
We start from the scaling laws~\cite{kaplan2020scaling}, which showed that the training of transformer language models is initially efficient with fast loss reduction, then the compute-efficient regime ends and the rate of the loss reduction slows down. 
In addition, the initial compute-efficient regime is longer for larger models. 
These ideas are illustrated in Figure~\ref{fig:optimal_schedule}\new{where $\trainingdynamics$ 
is initially large then it slows down}.
As shown in \citet{kaplan2020scaling} and \citet{li2020train}, the optimal compute allocation should favor a large model size and stop the training by the end of the initial compute-efficient regime
\new{when $\trainingdynamics = \optimalthreshold$, where $\optimalthreshold$ is some threshold. }
We call this training to ``Optimality'' as opposed to training to ``Completion'' or to convergence.
\new{We discuss later this section how to find the point of \optimality using constrained optimization in an idealistic scenario, 
then later in Section~\ref{sec:practical_schedule} using a more practical method that estimates $\optimalthreshold$.}


\paragraph{Intermediate Stages}
We next discuss where in training to grow a model.
Intuitively, the optimal schedule is one where the original small model is trained until its compute-efficient regime ends, then grown to a larger model to continue its compute-efficient regime.
Figure~\ref{fig:optimal_schedule} highlights one such potential schedule. 
Notice that there's a specific point on the loss curve of the original model that leads to the most compute saving; growing the model earlier means a wasted opportunity skipping some of the fast loss reduction stage of the original model, and growing the model later means wasting compute by continuing to train after the loss of the original model begins to plateau.

\paragraph{Schedule which minimizes compute}
Next, we describe how to mathematically find this optimal schedule.
For that, we use the scaling laws~\cite{kaplan2020scaling}  which derived empirical fits for the language model loss $L$ as it relates to compute $C$, number of non-embedding parameters $N$, number of gradient update steps $S$, and batch size $B$.  The total compute and the loss are given by
\begin{equation}
    \label{eq:scaling_laws}
    \begin{aligned}
    C  & \approx  6 N B S, \\
    L(N, S) &  =   \left( \frac{N_c}{N} \right) ^ {\alpha_N} + \left( \frac{S_c}{S} \right) ^ {\alpha_S}
    \end{aligned}
\end{equation}
where $\alpha_N, \alpha_S, \alpha_B, N_c, S_c, B_*$ are all model-specific constants. Thus, finding the the optimal schedule can be formulated as
a constrained optimization problem.
The output is the intermediate model sizes, and the amount of compute 
for each stage. We discuss the details in \appref{sec:optimal_schedule_appendix}.

\section{Practical Schedule}
\label{sec:practical_schedule}




While the general scaling laws are known, re-estimating their constants ($\alpha_N, \alpha_S, \alpha_B, N_c, S_c, B_*$ ) for our setup is challenging because it requires running a large number of models of different sizes. 
Instead of estimating all the constants, we make the observation that we only need to find three key points:
\new{
\begin{compactitem}
    \item \pregrowth $ \in \mathcal{L}(\model_{orig}, C)$ at which we grow the original model
    \item \growthtarget $\in \mathcal{L}(\model_{target}, C)$ that the model is grown towards to
    \item \optimality $ \in \mathcal{L}(\model_{grown}, C)$ at which we stop training the grown model.
\end{compactitem}
}
\new{
Next we discuss the mathematical definition of each point, how to find them, and how to use them in 
the actual training procedure. 

\paragraph{\pregrowth and \optimality points}

We define \pregrowth and \optimality using the slope of the loss curve, as they depend on the rate of change of the loss. Formally, 
\begin{equation}
\label{eq:thr}
\begin{aligned}
\pregrowth&: \frac{\partial }{\partial C}  \mathcal{L}(\model_{orig}, C) = \growththreshold\\
\optimality&: \frac{\partial }{\partial C}  \mathcal{L}(\model_{grown}, C) = \optimalthreshold
\end{aligned}
\end{equation}
where $\growththreshold$ and $\optimalthreshold$ are empirically estimated thresholds. 


Importantly, both thresholds are independent of the model size, and this independence can be derived from \eref{eq:scaling_laws}. We also empirically confirmed the model-size independence by training many models of different sizes; reconfirming results from \citet{kaplan2020scaling}, the shape of the loss curves for different sized models were similar, just shifted and scaled. 
Additionally, while both thresholds are model-size independent, $\growththreshold$ is a function of the growth operator. 

\paragraph{\growthtarget point}
The importance of the \growthtarget point is that it specifics the learning rate schedule of the grown model (\sref{sec:training_state}). We will specify  \growthtarget using the number of training 
steps ${S_\text{\growthtarget}}$. We found that it can be simply defined using
\begin{equation}
\label{eq:rho}
\begin{aligned}
{S_\text{\growthtarget}} = \rho \times S_\text{\pregrowth}
\end{aligned}
\end{equation}
where $\rho$ is an empirically estimated constant, and $S_\text{\pregrowth}$ is number of training 
steps at \pregrowth. 

As above, and knowing that the loss curves of models of different sized models are scaled and shifted versions of each other, we can use \eref{eq:scaling_laws} to show that $\rho$ is independent of the model size, and it is only a function of the growth operator. We also verified this empirically using different model sizes. 

\paragraph{Estimating $\tau$ and $\rho$}

Given that equations \ref{eq:thr} and \ref{eq:rho} are independent of model sizes, it is enough to estimate 
the values of $\tau$ and $\rho$ using small models. To estimate them, we first identify the three  
necessary points. 
Specifically, for a single growth operator, we train an original model and a target model from scratch then follow the intuition discussed in \sref{sec:optimal_schedule} to choose a \pregrowth point (on $\mathcal{L}(\model_{orig}, C)$) and an \optimality point (on $\mathcal{L}(\model_{target}, C))$. The \growthtarget point is simply the point on $\mathcal{L}(\model_{target}, C))$ with the same loss at \pregrowth. 
A plot like \fref{fig:optimal_schedule} (and the more detailed \fref{fig:growth_operator_compute_base}) make it easy to manually identify the three points, and we leave it to future work to automatically find these points.

Notice that this method required training a target model from scratch, but in practice we can estimate the constants once for smaller model sizes and apply them to larger sizes.
Using the identified points, we use equations~\ref{eq:thr}~and~\ref{eq:rho} to estimate values of $\growththreshold$ and $\rho$ for each growth operator, and the value of $\optimalthreshold$. The empirical values we estimated are in \appref{sec:practical_schedule_appendix}. Notice that estimating these constants is much simpler than estimating all the constants of the scaling laws 
($\alpha_N, \alpha_S, \alpha_B, N_c, S_c, B_*$) making this procedure simpler to apply to a new setup.
}

\new{

\begin{algorithm}[tb]
  \caption{Staged training for transformer LMs}
  \label{alg:training}
\begin{algorithmic}
\REQUIRE
  \STATE $\model$: original model
  \STATE $\lambda(t)$: learning rate schedule
  \STATE $M$: number of stages
  \STATE $(\growth_i, \tau_{_{\growth, i}}, \rho_{_{\growth, i}})$: $($growth op, $\tau$, $\rho)$ for stage $i$ 
  \STATE $\growth_1$: first stage operator assumed to be identity operator
  \STATE $\optimalthreshold$: last stage's $\tau$

\ENSURE
  \STATE $t \gets 0$   ~~~~~~~~~~~~~~~~~~~~~~~ // number of training steps
  \FOR{$i=1$ {\bfseries to} $M$}
   \IF{$i = M$} 
     \STATE $\tau \gets \optimalthreshold$ ~~~~~~  // last stage, stop at optimality
  \ELSE
     \STATE $\tau \gets \tau_{_{\growth, i}}$ 
  \ENDIF
  \WHILE{$\frac{\partial }{\partial C}  \mathcal{L}(\model, C) \leq \tau$}
  \STATE Run training step and update $\model$
  \STATE $t \gets t + 1$ ~~~~~  // update learning rate using $\lambda(t)$
 \ENDWHILE 
  \STATE $\model \gets \growth(\model)$
  \STATE $t \gets t \times \rho_{_{\growth, i}}$ ~~ // set learning rate for next stage
  \ENDFOR
    \STATE {\bfseries return} $\model$
\end{algorithmic}
\end{algorithm}

\paragraph{Training Procedure} 
Algorithm~\ref{alg:training} summarizes the staged training procedure. Notice that it extends it to the $M$ stage case. 
The algorithm starts with the model $\model$, trains it from scratch until  \pregrowth, grows the model, sets number of steps for the grown model, 
and repeats until no more stages, then continues training to \optimality.
}

\section{Experiments}
In this section we present our main empirical results, focusing on the amount of compute saving.
We show results on in-domain data (validation loss) and in the zero-shot transfer setting.
We also compare our work to prior work in growing transformer language models, establishing that previous methods fall short in one or more areas. 

\subsection{Experimental setting}
We experiment with GPT2 from \cite{radford2019language} (in base and large sizes) using the public C4~\cite{raffel2020exploring} dataset. 
We follow the learning rate schedule in \citet{kaplan2020scaling} for all model sizes, where the warmup period is set to 3,000 steps, the batch size is set to 512 and the sequence length is 1,024. 
For the zero-shot transfer learning, we experiment on two tasks: Wikitext-103~\cite{merity2016pointer} and LAMBADA~\cite{paperno2016lambada}, similar to \cite{li2021curriculum}. We report the compute saving for in-domain validation loss and zero-shot performance. 
We compare our $\text{practical}$ schedule in \sref{sec:practical_schedule} with the $\text{manual}$ schedule that directly matches the loss of the \pregrowth model with the target model to select the \growthtarget point.
We choose and report different thresholds to decide the \optimality of the training in each stage.

\subsection{Main results}
Figure~\ref{fig:optimal_schedule} shows the  compute saving for growing GPT2$_\textsc{large/4}$ original model to a target model of GPT2$_\textsc{large}$ (we show the results of other combinations of growth operators and the results of growing to GPT2$_\textsc{base}$ in~\appref{sec:additioal_result} in \fref{fig:growth_operator_compute_base}). 
It is clear that our grown models are reaching the same loss as the target models but with less compute. It is important to note that as both models train longer, the amount of compute saving drops. This illustrates a key design in our schedule where we stop
training of the grown model at \optimality when the compute-efficient regime ends.
Figure~\ref{fig:optimal_schedule} also demonstrates that small models achieve better loss trade-off when the total compute is limited, but saturate at higher loss when more compute is added, highlighting the advantages of our schedule in applying the growth operator to them before convergence.

\begin{table}[t]
\centering
\begin{small}
\resizebox{\columnwidth}{!}{
 \begin{tabular}{ll|rrr|rrr}
\toprule
 &   & \multicolumn{3}{c}{\gptlarge} & \multicolumn{3}{c}{\gptbase} \\ 
 &           & 5k       & 10k         & 14k     & 3.75k        & 7k      & 11k  \\
 \midrule
Baseline &  (\textbf{loss}) & 3.21  &  3.03   & 2.97  & 3.61  & 3.45  & 3.38 \\
\midrule
\midrule
\multicolumn{8}{c}{Compute savings (percent saved vs. baseline)}\\
\midrule
\multirow{2}{*}{1 stage$_\text{ manual}$} & 2xW &  19.3 & 5.6 & 4.0 & 23.8 & 14.5 & 13.8 \\
 & 2xD & 33.5 &  7.8 & 6.3 & 24.0 & 23.6 &21.8 \\
   \midrule
\multirow{5}{*}{1 stage$_\text{ practical}$} 
& 2xW & 22.5 & 7.3 & 5.2 & 24.3 & 20.2 & 19.7 \\
& 4xW & 18.0 & 5.3 & 3.8 & 16.0 & 8.6 & 5.5 \\
& 2xD & 37.0 & 11.0 &6.1 & 24.7 & 20.4 & 19.8  \\
& 4xD & 22.5 & 7.3 & 5.2 & 18.8 & 10.1 & 6.4 \\
& 2xDxW & 28.8 & 5.4 & 3.8 & 19.0 & 9.5 & 6.83 \\
          \midrule
\multirow{2}{*}{2 stage$_\text{ practical}$} & 2x2xW & 26.8 & 10.9 &  7.8 &  26.8 &  17.9 & 11.4 \\
 & 2x2xD & 30.0 & 14.5 & 10.4 & 33.3 & 21.4 & 15.9
 \\
 \bottomrule
\end{tabular}
}
\caption{Percentage compute savings for \gptlarge and \gptbase on in-domain validation loss on C4, ``W'' is width and ``D'' is depth growth operators. Percent savings is how much less compute our approach takes than the baseline to to train a model to equal or better loss than the baseline. Significant savings can be found using our both width and depth growth operators, and two growth stages can lead to even more savings than one growth stage. The derivative threshold at 5k steps is -0.1, at 10k steps it's -0.05, and at 14k steps it's -0.04; thus, 5k is undertrained, 10k is approximately at the optimality threshold of -0.052, and 14k is trained beyond optimality.  Similar for \gptbase.}
\label{tab:indomain}
\end{small}
\end{table}

\begin{table*}[ht]
\centering
\small
\resizebox{0.95\textwidth}{!}{
\begin{tabular}{ll|rrr|rrr|rrr|rrr}
\toprule
 &   & \multicolumn{6}{c}{Zero-shot Wikitext-103 (PPL)} & \multicolumn{6}{c}{Zero-shot LAMBADA (accuracy)} \\ 
 \midrule
 &   & \multicolumn{3}{c}{\gptlarge} & \multicolumn{3}{c}{\gptbase} & \multicolumn{3}{c}{\gptlarge} & \multicolumn{3}{c}{\gptbase} \\ 
 &   & 5k & 10k & 14k                & 3.75k        & 7k        & 11k& 5k & 10k & 14k                & 3.75k        & 7k        & 11k  \\  \midrule
Baseline  &  & 41.0  &  32.3  & 30.3 & 68.5   & 57.1  & 50.0 &  39.6 &  43.2   & 44.7  & 31.1 & 33.0  & 34.7  \\ \midrule \midrule
\multicolumn{14}{c}{Compute savings (percent saved vs. baseline, negative means more compute than baseline)}\\
\midrule
\multirow{2}{*}{1 stage$_\text{ manual}$} & 2xwidth &  -18.5 &  6.2 &  5.8  & -19.7 &  0.3 &  2.4 & 3.2 & 6.7 &  8.3 &  -8.3 & 0.2 & 13.6\\
& 2xdepth & 28.5 &  11.8 &  12.0 & 24.0 &  28.0 &  17.3  & 33.5 & 16.8 &  13.8 & 24.0 &  22.9 & 18.2\\
                       \midrule
\multirow{5}{*}{1 stage$_\text{ practical}$} 
& 2xwidth &  -20.5 & -0.25 &  8.7 &  -15.7 &   5.9 &  3.8 & -15.5 &  12.3 &  14.8 &  -15.7 &3.3 &  10.6\\ 
& 4xwidth &  -13.4 &  1.3 &  7.4 & -12.0 &  1.1 &  6.4 & -11.0 & 18.0 &  18.2 & -12.0 & 10.1 &  8.6\\ 
& 2xdepth &  32.0 &  13.5 &  11.4&  31.3 &  33.5 &  17.5 &  32.0 &  23.5 &  21.4 & 24.7 &  16.8 &  13.9\\
& 4xdepth &  21.0 &  17.5 &  9.5 &  14.6 &  7.9  &  3.1 & 21.0 & 20.5 &  14.6 & 14.6 &  9.4  & 7.8 \\ 
& 2xdepthxwidth &  -10.5 & -0.3 &  1.6 &  -12.0  &  3.6 &  4.5 & -95.0 & -37.5 &  0.7 & 5.4 & 6.4  &  6.4\\ 
        \midrule
\multirow{2}{*}{2 stage$_\text{ practical}$} & 2x2xwidth &  -2.5 &  6.3  &  9.3 &  3.3 &  5.4 &   8.0 & -3.3 & 13.4 &  20.3 &  -3.3 &  1.8 &  11.8 \\
 & 2x2xdepth &  30.0  &   12.5 &   12.8 &  33.3 & 14.3 &  11.8  & 19.0 & 14.5  &  23.6 &  19.9 & 10.6 &  15.0     \\  
 \bottomrule
\end{tabular}
}
\caption{Percentage compute savings for \gptlarge and \gptbase on out-of-domain Wikitext-103 (perplexity) and LAMBADA (accuracy). Percent savings is how much less compute our approach takes than the baseline to to train a model to equal or better perplexity or accuracy than the baseline; negative numbers mean our approach took more compute than the baseline to achieve equal or better performance. Results are mixed in the early stages of training, but our approach leads to compute savings for all experiments later in training (14k for large, 11k for base). The derivative threshold at 5k steps is -0.1, at 10k steps it's -0.05, and at 14k steps it's -0.04; thus, 5k is undertrained, 10k is approximately at the optimality threshold of -0.052, and 14k is trained beyond optimality. Similar for \gptbase.}
\label{tab:wikitext_and_lambada}
\end{table*}


Table~\ref{tab:indomain} shows the amount of compute saving of our growth operators for two different model sizes. The table shows that our grown model reaches the same performance of the baseline (target model) trained from scratch while having considerable compute saving ranging from 30\% to 5\% for different thresholds. 
The first row in Table~\ref{tab:indomain} denotes the number of steps we used to train the baseline model.
Given that our growth operator is loss-preserving and training-dynamic preserving, we can always reach the same loss of the
target model
with less compute and the compute saving becomes larger when we decide to stop the target model earlier.
Also, given the same growth ratio (4x growth), the depth growth operator is preferable versus width concerning the compute saving. 

We also evaluate our pretrained models on other language modeling tasks in the zero-shot setting to verify that the grown models maintain their transfer learning capabilities. 
Table~\ref{tab:wikitext_and_lambada} shows results on Wikitext-103 and LAMBADA.
It can be seen that using our practical schedule, we achieve comparable and sometimes better performance versus using the manual schedule for both in-domain loss and zero-shot transfer learning. 
In some cases we have negative compute saving with the width operator or when combing the width and depth operator on zero-shot transfer learning tasks early in training (indicating the grown model used more compute to get to the same (or better) performance), but the compute saving is always positive when training for longer. 
We assume that this is due to instabilities in optimization right after applying the growth operator. 
When the training proceeds, the zero-shot performance will shortly recover and the grown models will have better positive compute saving at \optimality for the two model sizes. 
It is also less of an issue for the \texttt{base} model size as \growthtarget. 
Moreover, the depth growth operator leads to better performance compute saving trade-off compared to the width operator under the same growth ratio.  
Finally, we show that applying the growth operator twice (in two stages) lead to the best-performing compute saving and performance.  
See~\fref{fig:wiki_lambada} for detailed evaluation plots.

\subsection{Ablations and prior work comparison}
We experiment with prior work and conduct ablation studies for our method. We show that 
prior works fall short in one or more of the key components of our proposed method; preserving loss, preserving training dynamics, or following an optimal schedule. 

\begin{figure*}[t]
    \centering
    \begin{subfigure}{.22\textwidth}
        \centering
        \includegraphics[width=\textwidth]{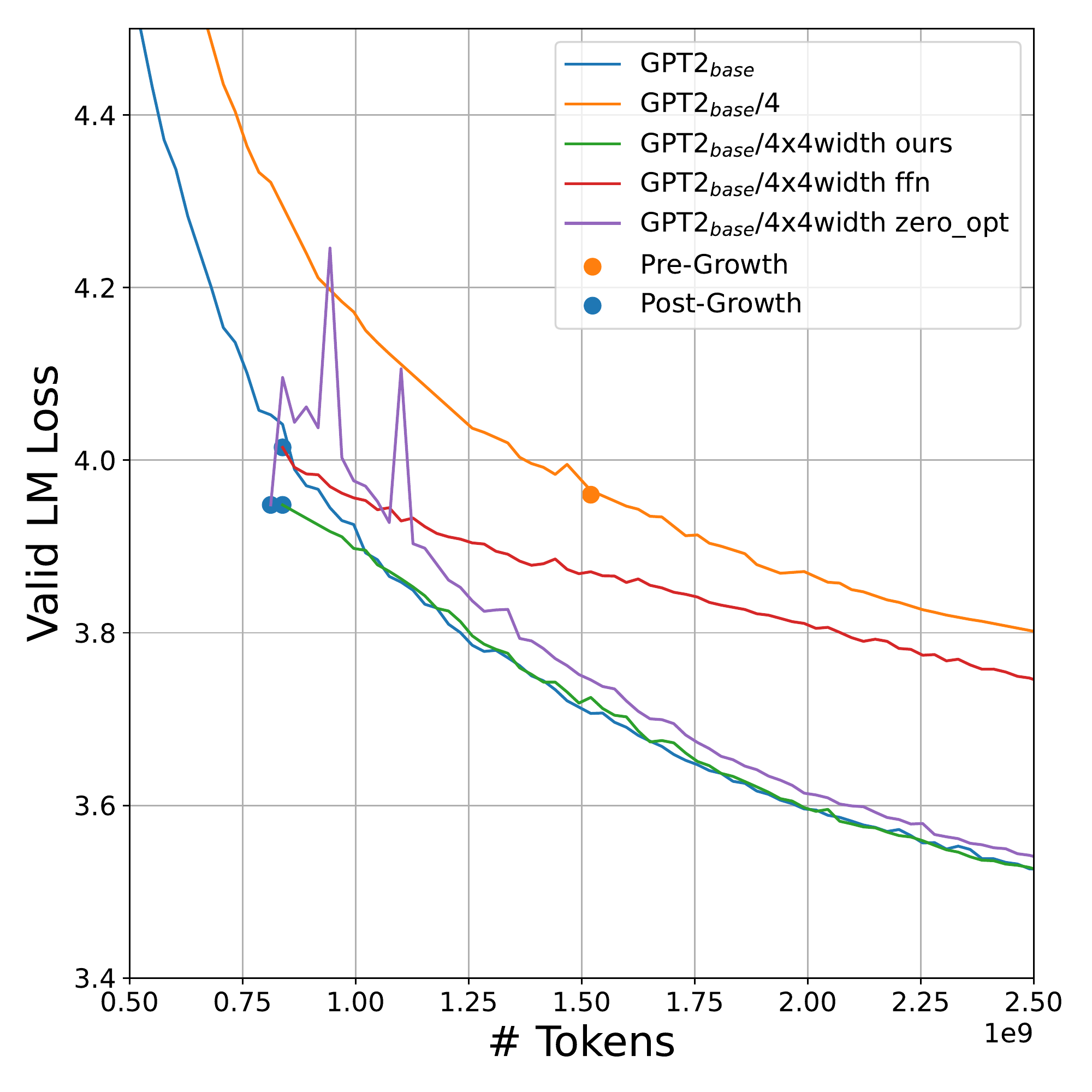}
        \caption{
        Width op. and optimizer}
    \end{subfigure}%
    \begin{subfigure}{.22\textwidth}
        \centering
        \includegraphics[width=\textwidth]{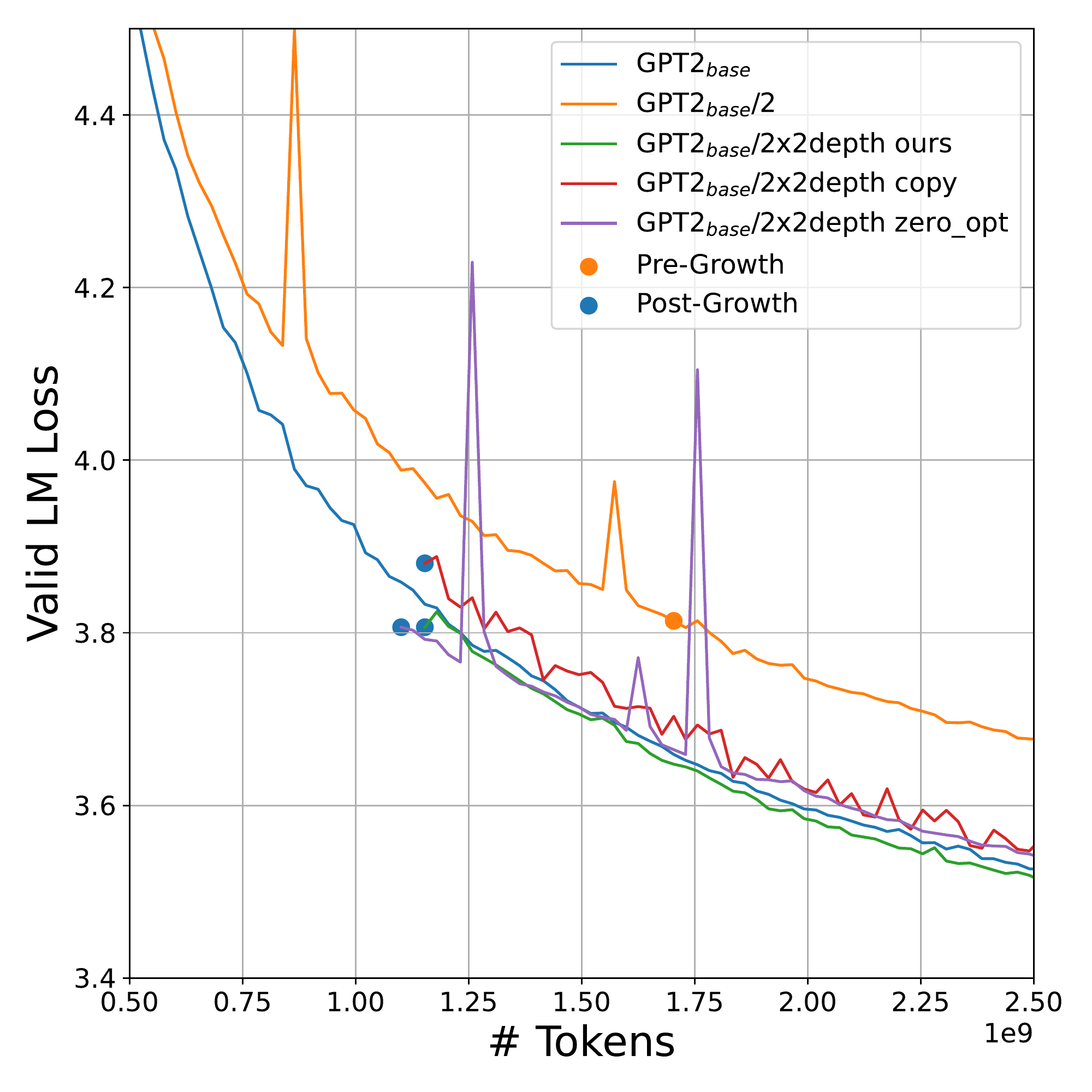}
        \caption{Depth op. and optimizer }
    \end{subfigure}
    \begin{subfigure}{.22\textwidth}
        \centering
        \includegraphics[width=\textwidth]{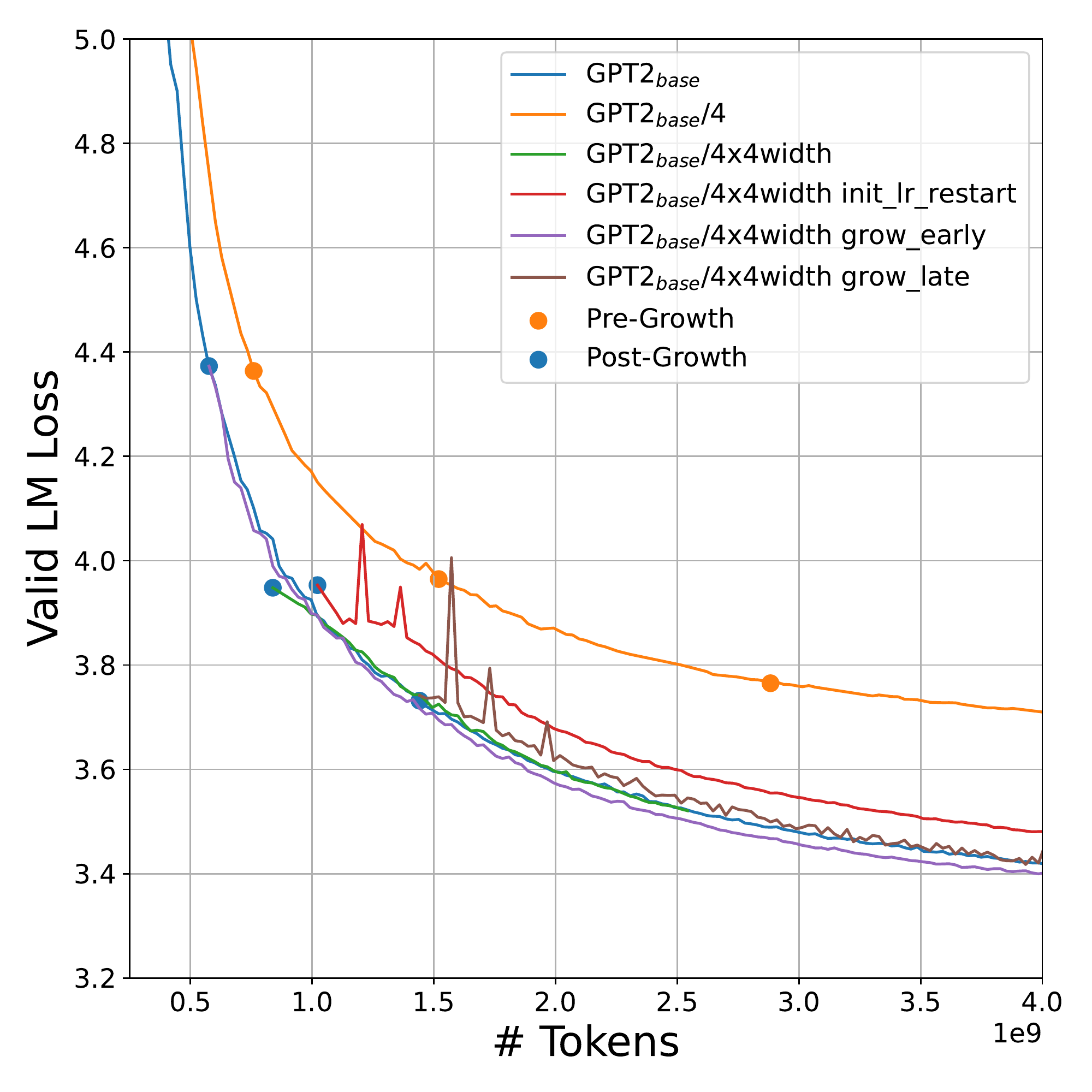}
        \caption{LR and optimal schedule}
        \label{subfig:grow_earlylate_tokens}
    \end{subfigure}%
    \begin{subfigure}{.22\textwidth}
        \centering
        \includegraphics[width=\textwidth]{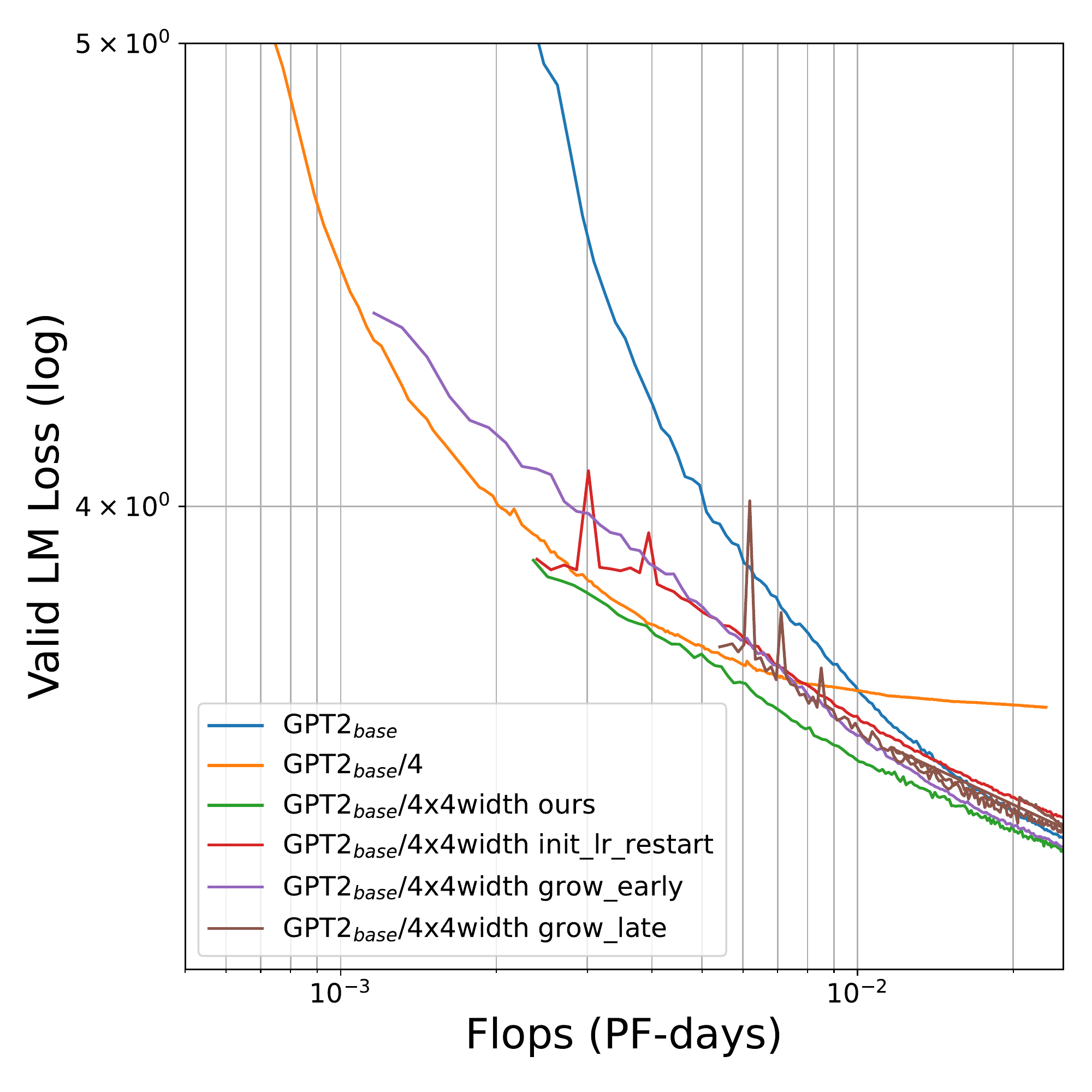}
        \caption{LR and optimal schedule}
        \label{subfig:grow_earlylate_compute}
    \end{subfigure}%
    \caption{
    Comparing with three different baselines from prior work and ablation studies. 
    $\bullet$~The width growth operator of \citet{gu2021transformer} in {GPT2${_\text{\textsc{base}/4x4width ffn}}$} and the depth growth operator
    of \citet{gong2019efficient} in {GPT2$_\text{\textsc{base}/2x2depth copy}$}
    are not loss preserving (higher initial loss). Also,  {GPT2${_\text{\textsc{base}/4x4width ffn}}$} is significantly underperforming the target model GPT2$_\textsc{base}$. $\bullet$~Resetting the optimizer state to zero instead of growing it (the \texttt{zero\_opt} runs) have large instabilities and not preserving of the training dynamics. 
    $\bullet$~\ref{subfig:grow_earlylate_tokens}
    shows that restarting the learning rate schedule as in ~\citet{rae2021scaling} is not training-dynamics-preserving.
    $\bullet$~\ref{subfig:grow_earlylate_tokens}, \ref{subfig:grow_earlylate_compute} also show that 
    not following our optimal schedule and grow the model too early or too late is still loss-preserving and training-dynamics-preserving but leads to lower compute saving
    }
    \label{fig:growth_operator}
\end{figure*}

\paragraph{Prior work comparison.}
We evaluate against two growth operators from previous work \citep{gu2021transformer,gong2019efficient}; one uses weight sharing to make a feed-forward network module wider, and the other makes an entire network deeper.
Given the nature of these growth operators, the grown models do not represent the same function as the original model; as shown in \fref{fig:growth_operator}, neither retain the same loss as the original model, and thus they do not satisfy our loss-preserving property.

Prior work also mostly ignored the optimizer state. In ~\fref{fig:growth_operator}, we explore this as an ablation by simply setting the optimizer state to zero for our width and depth growth operators.
Though the loss can be retained at the starting pointing for the grown model, the training dynamic becomes extreme unstable after applying the growth operators, and thus such a growth operator will not have the training-dynamics-preserving property.

\paragraph{Ablation studies}
We further perform two ablation studies concerning the learning rate schedule and growth schedule in ~\fref{fig:growth_operator}. 
For the learning rate schedule, we compare our setting of the learning rate to the \growthtarget point with restarting the learning rate schedule as in~\citet{rae2021scaling}. 
It can be seen that resetting the learning rate schedule leads to much unstable training dynamics that are not  aligned with the \growthtarget. 
For the growth schedule, we experiment with growing the small model earlier or later than the \growthtarget point computed with our proposed practical schedule. 
It shows that growing the model too late/early is still loss-preserving and training-dynamics-preserving but 
the grown models lose the compute saving advantages. 

\paragraph{Training to completion vs. to \optimality.}
While training to \optimality is not one of our contributions, it is an important part of our training schedule. 
Here we compare our models in \tref{tab:wikitext_and_lambada} with equivalent models trained to completion. 
In~\citet{radford2019language} , GPT2$_\textsc{base}$ trained to completion achieves 37.5 PPL on Wikitext-103 and 45.9 accuracy on LAMBADA. 
Our best-performing 2x2xdepth grown GPT2$_\textsc{large}$ model achieves the same PPL on Wikitext-103 as this model with only 15\% of the compute 
and the same accuracy as the this model on LAMBADA with 33\% of the compute. 
Notice that we are comparing different model sizes,  a large model trained to \optimality vs. a smaller model trained to completion.





\section{Related Work}
\label{sec:related}


%

Perhaps the most similar prior work is \cite{gu2021transformer}.
However, this work did not provide a method to decide when to apply a growth operator and instead evaluated the performance of their operators at 100/300/500/700K steps of a small model, or applied a heuristic to equally distribute training steps among different model sizes. They did not discuss the optimizer state, and they reset the learning rate to the maximum value of 1e-4 at the beginning of each state without warmup. Their work built on progressive stacking (discussed below), and their proposed method to grow the width only grew the feed-forward layers instead of the entire model width.

\cite{gong2019efficient} proposed ``progressive stacking'' which doubles the depth of a BERT transformer model by copying layers; to construct a 2$L$-layer model from a $L$ layer model it copies layer $i \leq L$ in the smaller model to layer $(i + L)$ in the larger model. 
The optimizer state is reset at the beginning of each stage, but the learning rate is kept the same as the prior stage. They use heuristics to set the stacking schedule: 50K steps for 3-layer, 70K steps for 6-layer model, 280K steps for 12-layer model. 
In an ablation study they examined the sensitivity to the number of steps before applying their growth operator and concluded that there is a threshold number of steps and for switching times such that switching to the larger model before the threshold led to compute savings, but switching after the threshold didn't.  This is consistent with our results that showed the amount of compute saving is closely related to the stage length.

\cite{li-etal-2020-shallow} proposed growing encoder-decoder transformers in the context of training a machine translation system.  Their depth growth operator is identical to progressive stacking, although they explore operations that increase the model by only copying some of the layers from the small to large model (e.g. growing from 12 to 18 layers). 
They do not mention optimizer state, and reset the learning rate to max value at each stage.

In contemporaneous work, Gopher \cite{rae2021scaling} introduced a method which tiles the weights from a small model to a larger one. However, their growth operator does not satisfy our two properties from Section~\ref{sec:properties}, and they focus on training their models to completion.
~\citet{evci2022gradmax} proposes a way to initialize the grown weights by maximizing the gradient norm of the new weights for vision models. Their growth operator also requires specified activation functions that can not be directly applied to transformers.
Finally, \citet{li2021curriculum} proposed applying a curriculum learning strategy to the sequence length to reduce the training cost when training large language models. 
Their work is orthogonal to our method and our methods could be combined; we leave this to future work.

\section{Conclusion and future work}
One direction of future work is to combine batch size warmup~\cite{brown2020language} and sequence length growth~\cite{li2021curriculum} with our depth and width growth operators.
Another applies our proposed methods to train a massive transformer.

We presented a staged training method for large transformer-based language models that grows the model size during training.
We demonstrated the importance of two properties of the growth operators (loss-preserving and training-dynamics-preserving), and provided depth and width operators that satisfy both requirements.
Finally, we devised a principled approach to find the optimal schedule and  a simple method to apply the schedule in practice. Empirical evaluations show up to 22\% compute saving.


\bibliography{acl_2022}
\bibliographystyle{acl_natbib}

\appendix

\section{Additional Plots and Results}
\label{sec:additioal_result}
\begin{figure*}[t]
    \centering
    \begin{subfigure}{.45\textwidth}
        \centering
        \includegraphics[width=\textwidth]{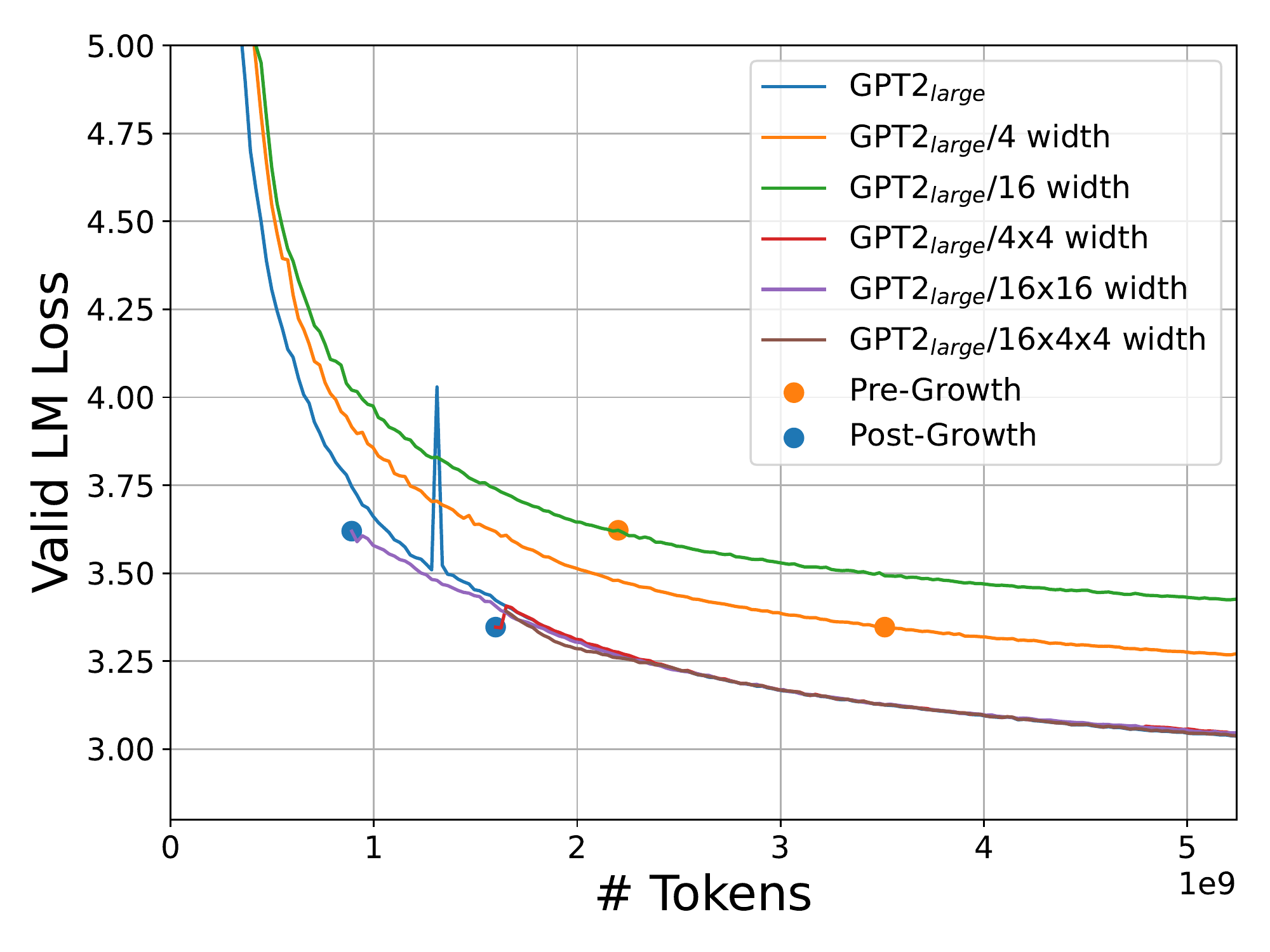}
        \caption{Width growth with GPT2$_\textsc{large}$ as target}
    \end{subfigure}
    \begin{subfigure}{.45\textwidth}
        \centering
        \includegraphics[width=\textwidth]{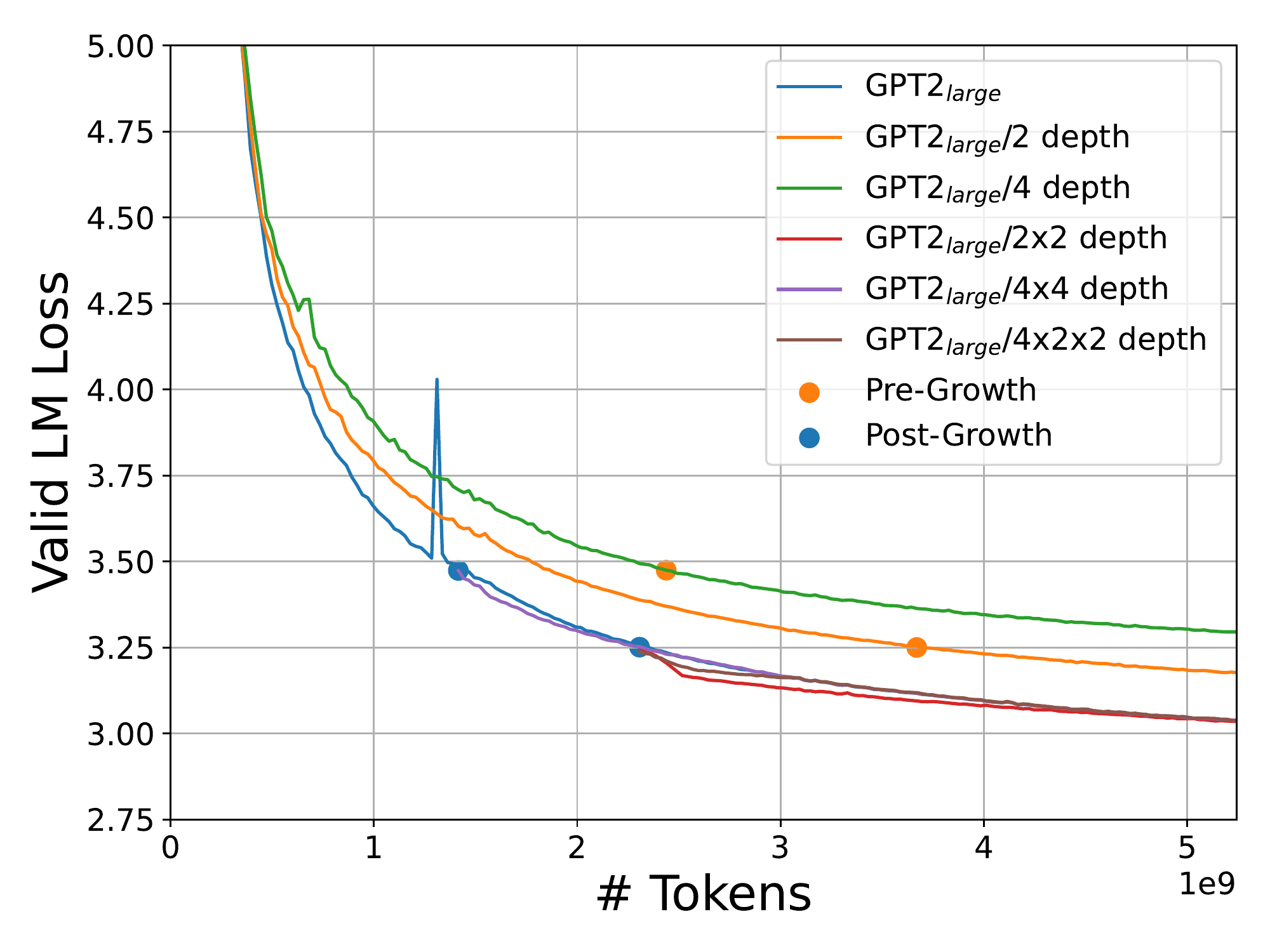}
        \caption{Depth growth with GPT2$_\textsc{large}$ as target}
    \end{subfigure}
    \caption{Similar to \fref{fig:width_depth_token}, our width and depth growth operators are loss-preserving and training dynamics preserving. 
    $\bullet$~{GPT2$_\text{\textsc{large}/16x16-\textsc{width}}$} indicates  starting from a 16x smaller model then growing it 16x by doubling the width twice. 
    $\bullet$~{GPT2$_\text{\textsc{large}/16x4x4-\textsc{width}}$} indicates growing the model 16x over two stages, by doubling the width once, continue training the model, then doubling the width again.
    $\bullet$~The same applies to the depth growth operator.
    }
    \label{fig:width_depth_token_base}
\end{figure*}

\begin{figure*}[t]
    \centering
    \begin{subfigure}{.45\textwidth}
        \centering
        \includegraphics[width=\textwidth]{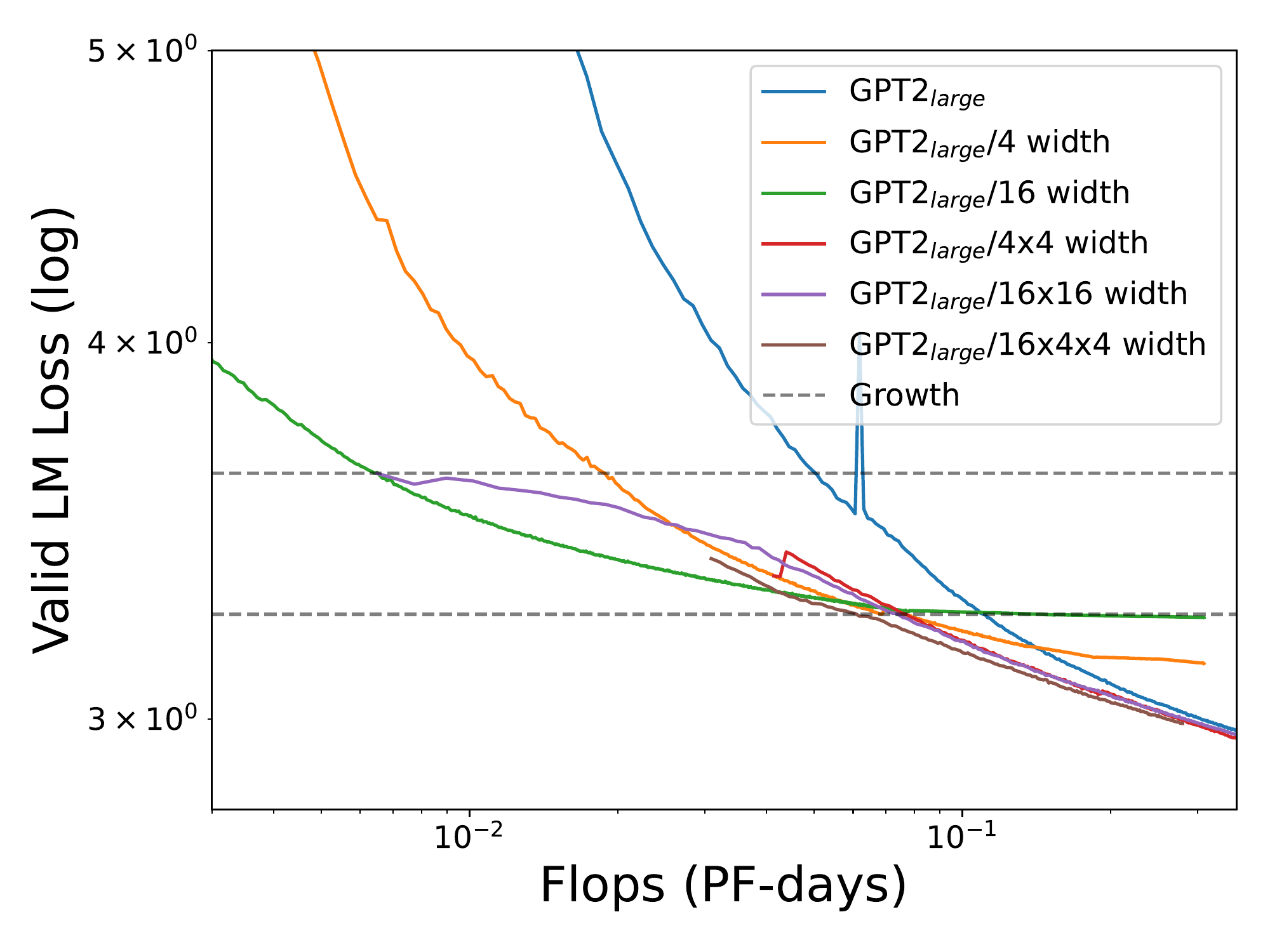}
        \caption{Width growth with GPT2$_\textsc{large}$ as target}
    \end{subfigure}%
    \begin{subfigure}{.45\textwidth}
        \centering
        \includegraphics[width=\textwidth]{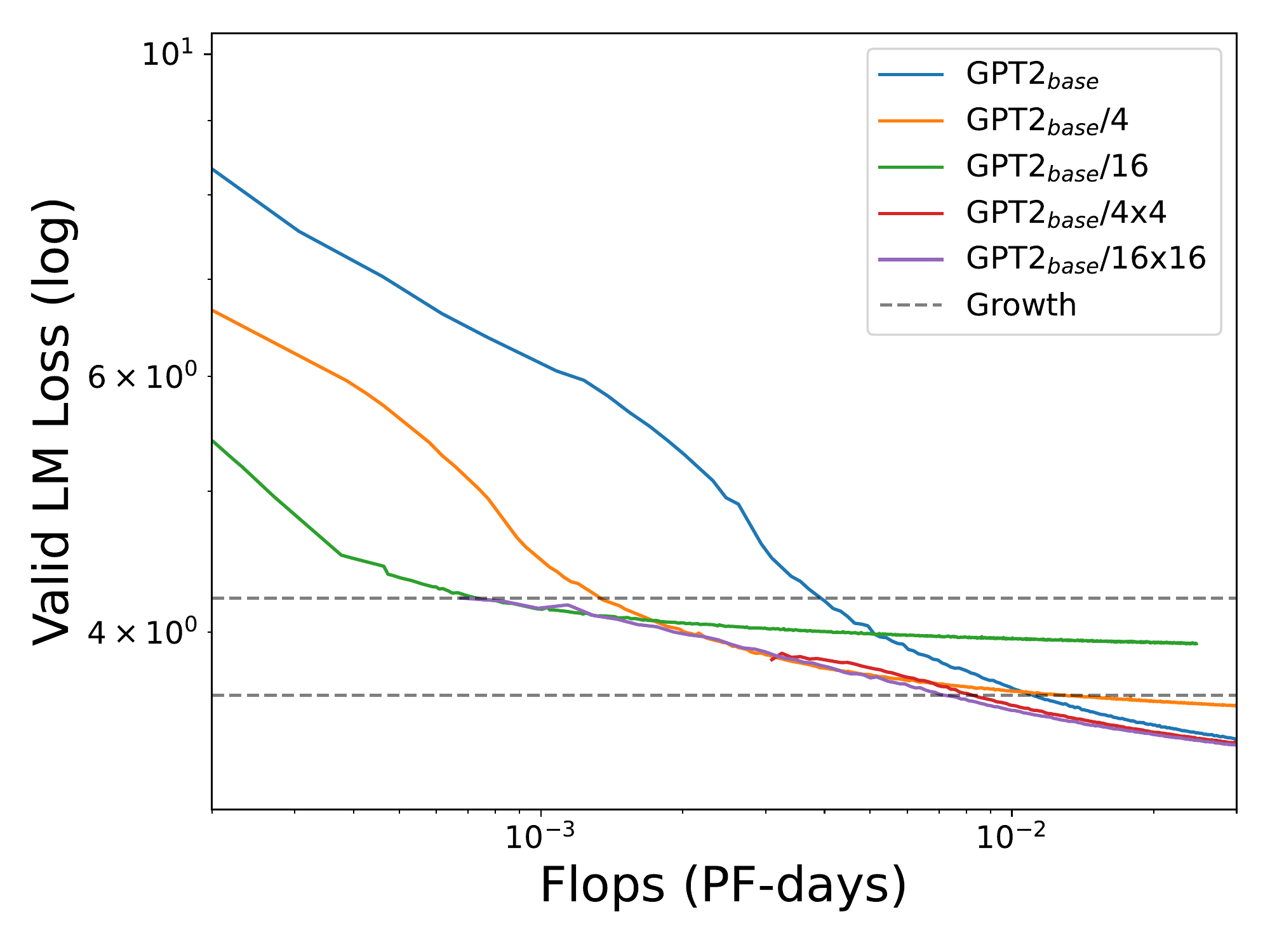}
        \caption{Width growth with GPT2$_\textsc{base}$ as target}
    \end{subfigure}
    \begin{subfigure}{.45\textwidth}
        \centering
        \includegraphics[width=\textwidth]{figs/figure_gpt2base_width_compute.pdf}
        \caption{Depth growth with GPT2$_\textsc{large}$ as target}
    \end{subfigure}
    \begin{subfigure}{.45\textwidth}
        \centering
        \includegraphics[width=\textwidth]{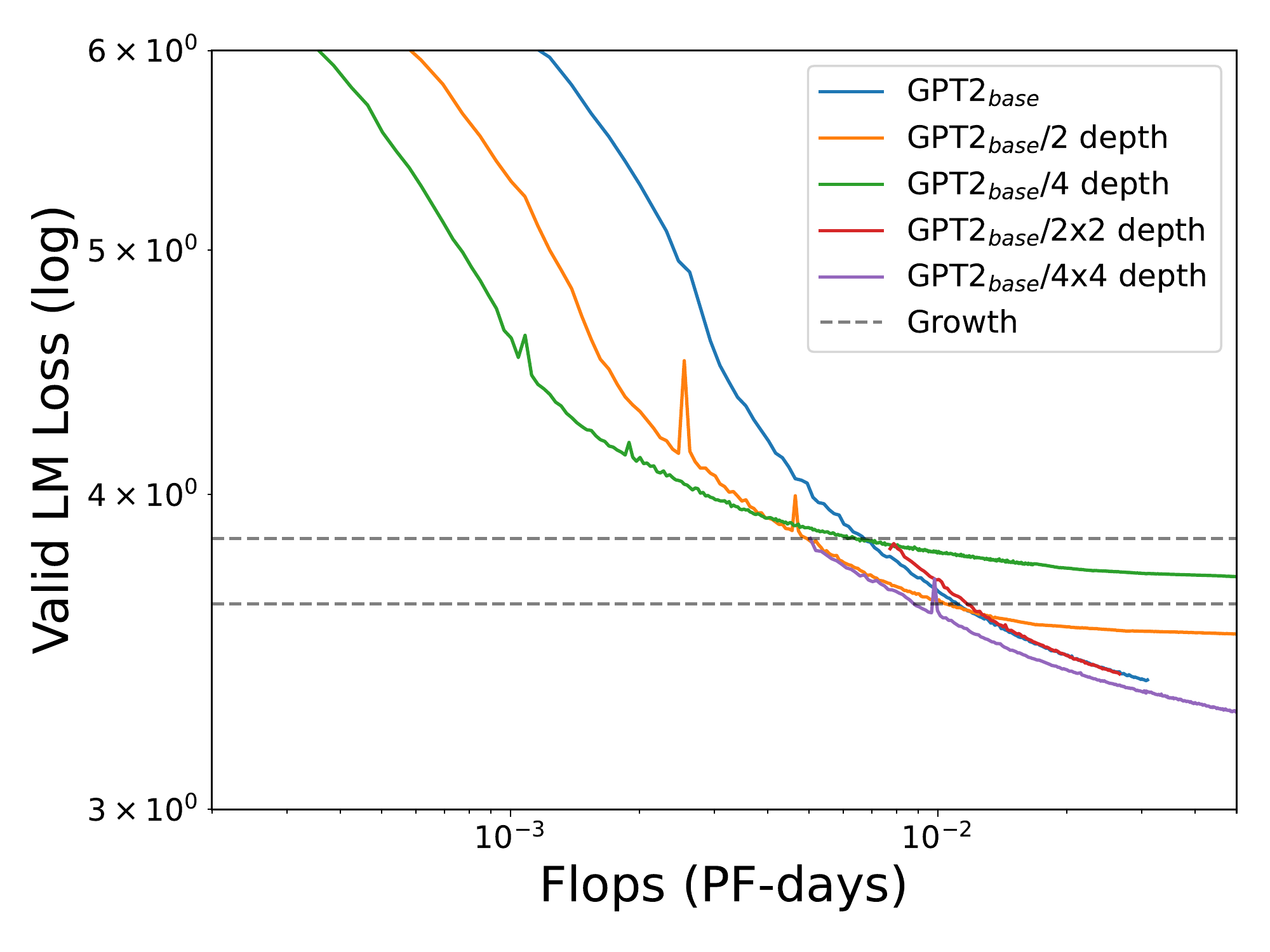}
        \caption{Depth growth with GPT2$_\textsc{base}$ as target}
    \end{subfigure}
    \caption{Growth operator and total compute. Our grown models are saving compute compared with target model. 
    }
    \label{fig:growth_operator_compute_base}
\end{figure*}

In~\fref{fig:width_depth_token_base} and \ref{fig:growth_operator_compute_base}, we show loss curve plot and compute plot for applying growth operator to GPT2$_\textsc{base}$ model.
These suggest our growth operator is loss-preserving,  training dynamic preserving and saves compute to train the target \textsc{base} size model. 

\begin{figure*}[t]
    \centering
    \begin{subfigure}{.65\textwidth}
        \centering
        \includegraphics[width=\textwidth]{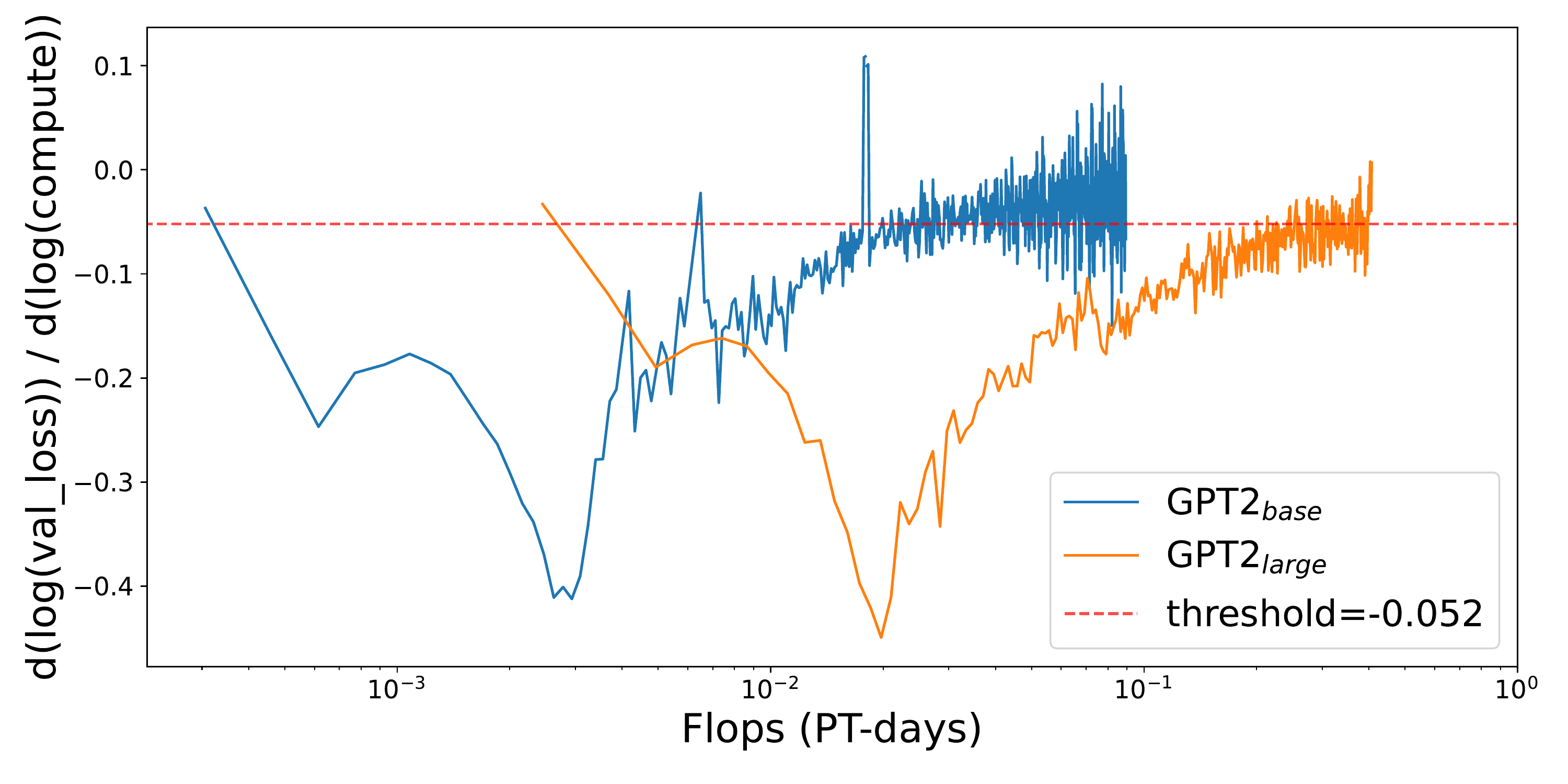}
        \caption{First derivative on the validation loss curve for GPT2$_\textsc{base}$ and GPT2$_\textsc{large}$. This is an approximation to the training dynamics $\trainingdynamics$.}
        \label{subfig:first_derivative}
    \end{subfigure}
    \begin{subfigure}{.65\textwidth}
        \centering
        \includegraphics[width=\textwidth]{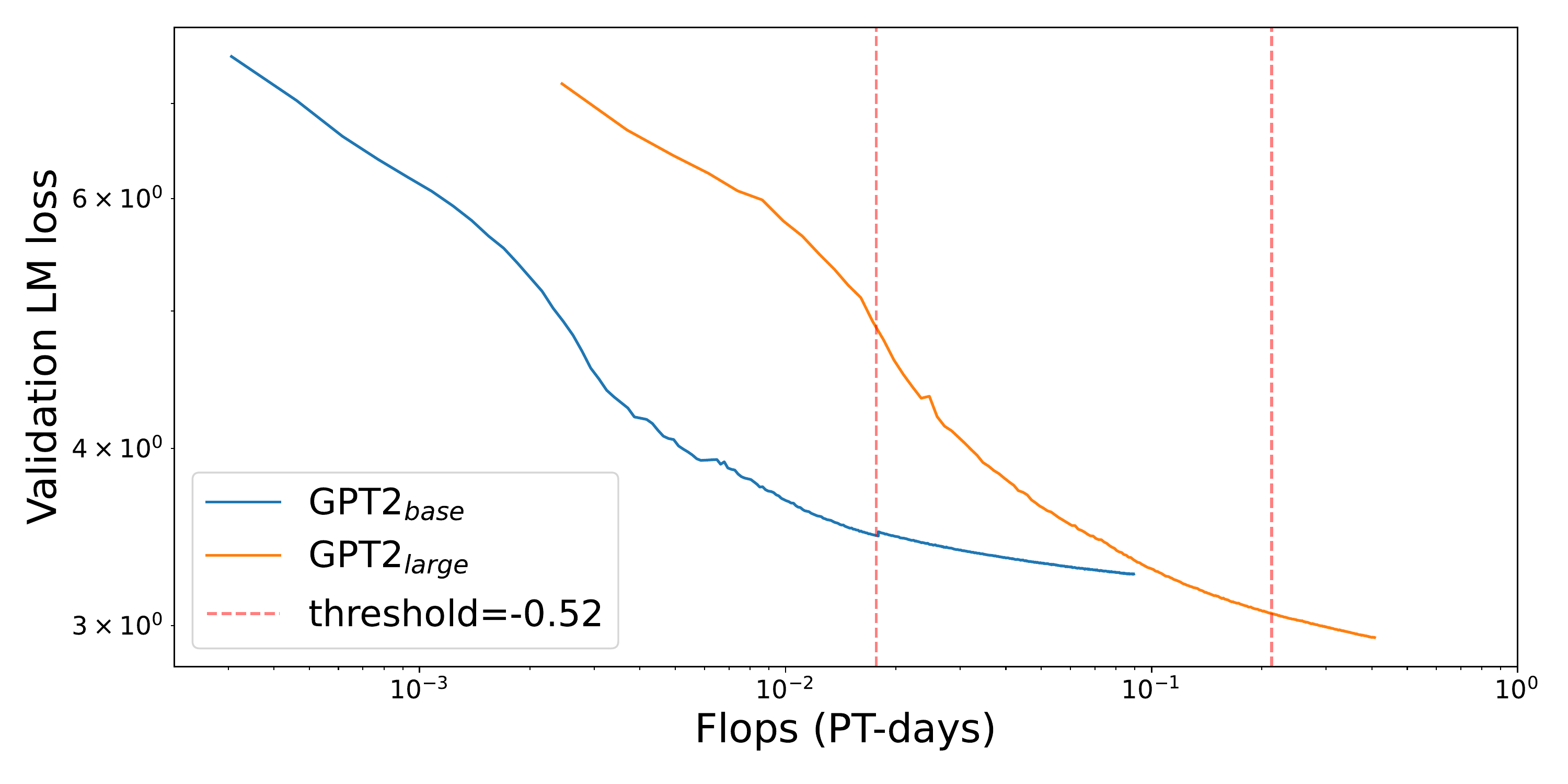}
        \caption{Validation loss curve for GPT2$_\textsc{base}$ and GPT2$_\textsc{large}$}
        \label{subfig:valid_loss}
    \end{subfigure}
    \caption{
    ~\ref{subfig:first_derivative} shows the first derivative of the validation loss curve for GPT2$_\textsc{base}$ and GPT2$_\textsc{large}$. 
    The horizontal dotted lines shows the empirical value of the derivative threshold we use to identify the point of \optimality. 
    ~\ref{subfig:valid_loss} shows how the threshold value in ~\ref{subfig:first_derivative} maps to a point in the loss curve for both model sizes (the first vertical line for GPT2$_\textsc{base}$ and the second for GPT2$_\textsc{large}$). Both point show slowed training and the end
    of the compute-efficient regime.  
    }
    \label{fig:first_derivate}
\end{figure*}

In~\fref{fig:first_derivate}, we demonstrate the loss curve and derivatives regarding the validation loss and compute in log scale. It clearly shows the used practical threshold gives us a good estimate of the \optimality of the loss curve. 

\begin{figure*}[h]
    \centering
    \begin{subfigure}{.45\textwidth}
        \centering
        \includegraphics[width=\textwidth]{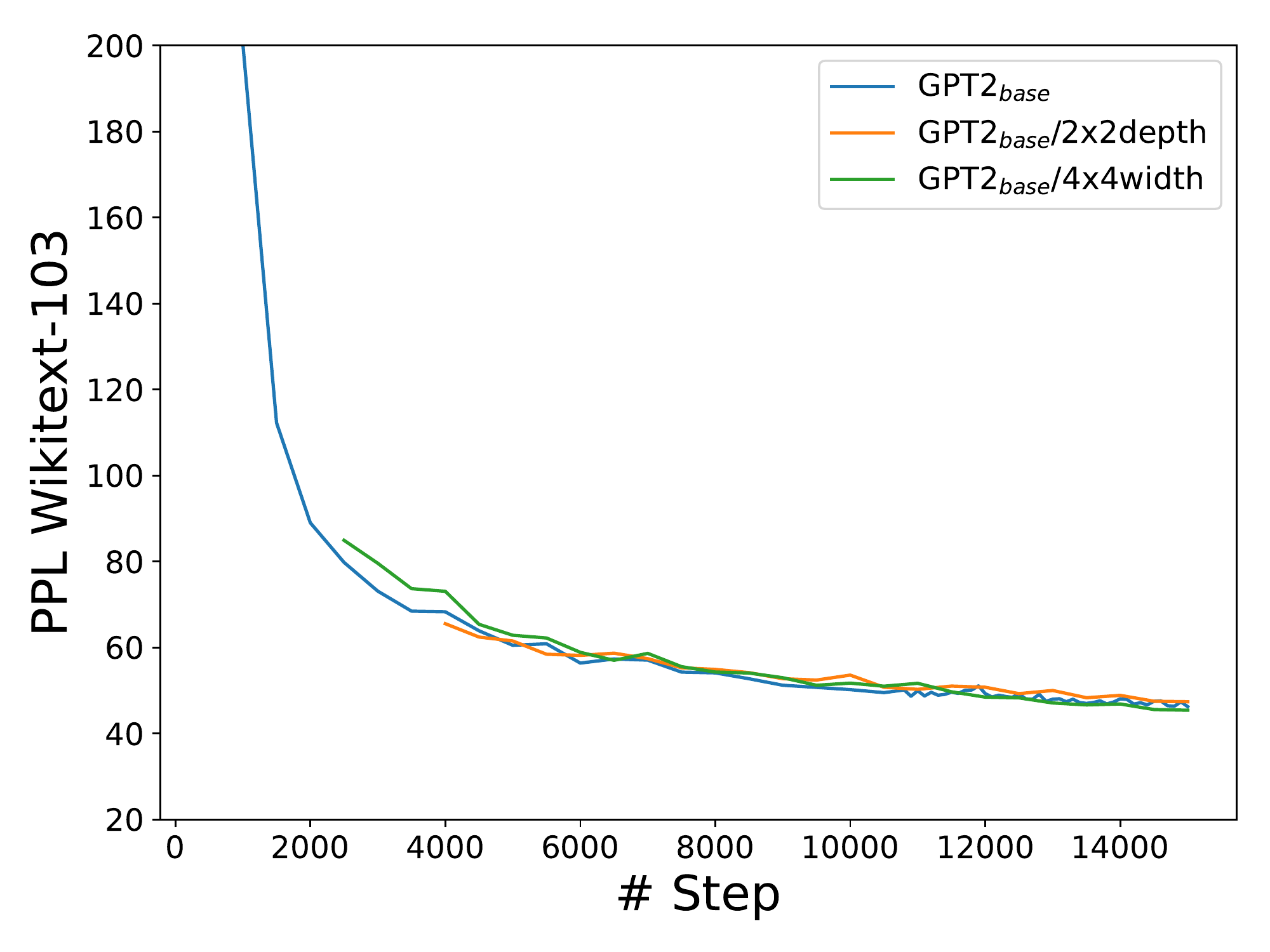}
        \caption{Wikitext-103 evaluation with GPT2$_\textsc{base}$ as target}
    \end{subfigure}
    \begin{subfigure}{.45\textwidth}
        \centering
        \includegraphics[width=\textwidth]{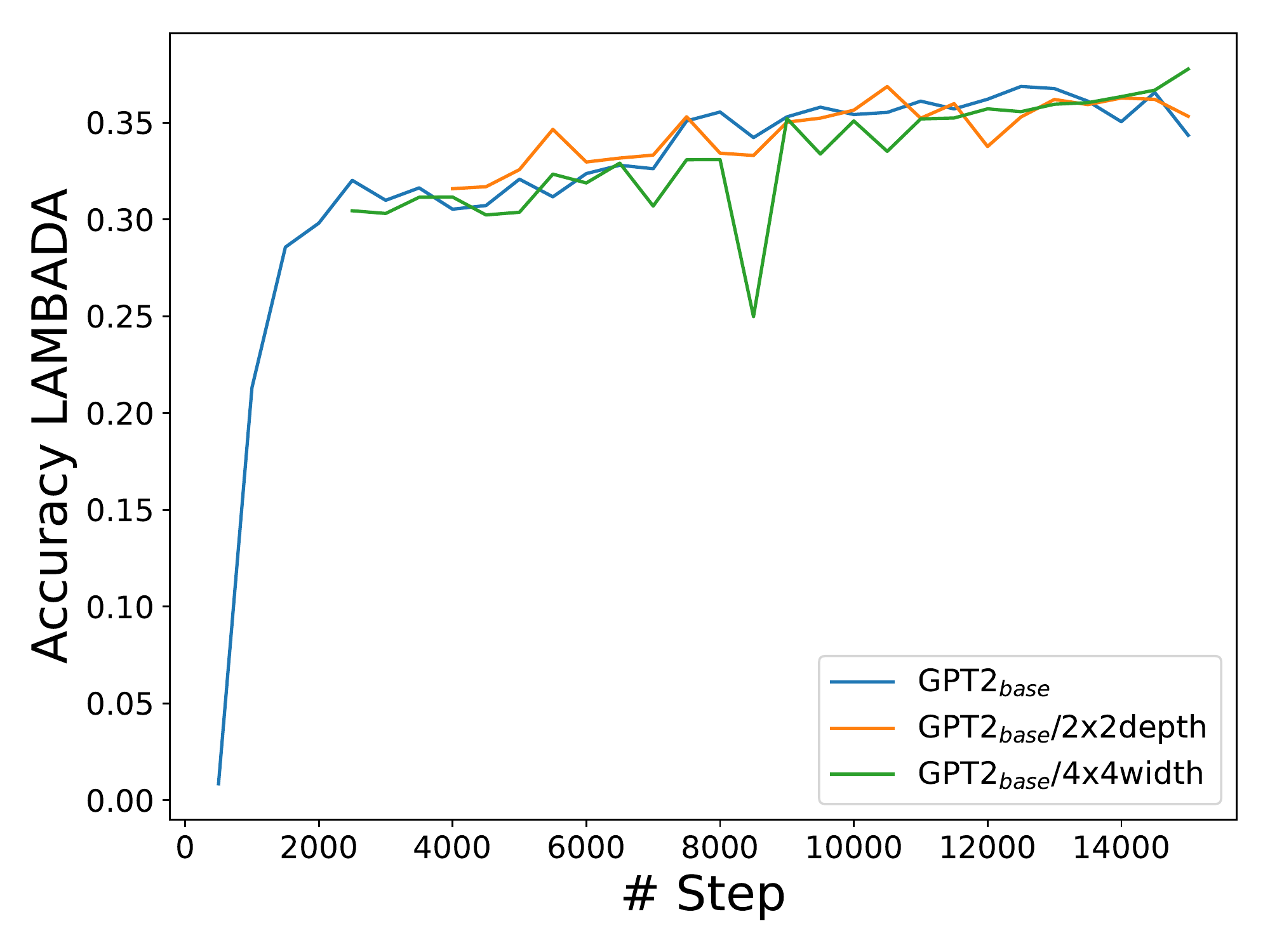}
        \caption{LAMBADA evaluation with GPT2$\textsc{base}$ as target.}
    \end{subfigure}
    \caption{The detailed evaluation plots on Wikitext-103 and LAMBADA for GPT2$_\textsc{base}$ trained from scratch and two grown model using depth or width operator. The performance of the GPT2$_\text{\textsc{base}/2x2depth}$ perfectly follows or even outperform the target model, while GPT2$_\text{\textsc{base}/4x4width}$ underperforms the target model initially after growing but it  catches up quickly. This explains the initial ``negative'' compute saving  in \tref{tab:wikitext_and_lambada} for the width growth operator, followed by positive compute saving as the model trains longer. 
    }
    \label{fig:wiki_lambada}
\end{figure*}

In~\fref{fig:wiki_lambada}, we present the evaluation results for every checkpoints across the steps for target and grown GPT2$_\textsc{base}$ model. The performance of the GPT2$_\text{\textsc{base}/2x2depth}$ can always perform on par with or better than the baseline while GPT2$_\text{\textsc{base}/4x4width}$ underperforms baseline in the initial phase after growing but can catch up quickly. This also explains the potential negative compute we have in the main text in the early phase of the evaluation and better compute saving at \optimality. 

\section{Model Architecture}
We use GPT2$_\textsc{base}$ and GPT2$_\textsc{large}$ models. For GPT2$_\textsc{base}$ model, it consists of 12 layers, 768 hidden dimensions, 12 heads and 125M parameters. For GPT2$_\textsc{large}$ model, it consists of 24 layers, 1536 hidden dimensions, 16 heads and 760M parameters. 

\section{Optimal Stage Schedule}
\label{sec:optimal_schedule_appendix}
This appendix defines a constrained optimization problem that produces the optimal staged-training schedule. 
The scaling laws of \citet{kaplan2020scaling} derived empirical fits for the language model loss $L$ as it relates to the total amount of compute $C$, number of non-embedding parameters $N$, number of gradient update steps $S$, and batch size $B$.  
The total compute\footnote{This neglects contributions proportional to the context length, $n_{ctx}$, and may not be valid in regime of large $n_{ctx}$ where $n_{ctx} \ge 12 d_{model}$} is given by
\begin{equation}
    C \approx 6 N B S,
\end{equation}
and the loss $L$ for any model size $N$ and number of steps $S$ is given by: 
\begin{equation}
    L(N, S) = \left( \frac{N_c}{N} \right) ^ {\alpha_N} + \left( \frac{S_c}{S} \right) ^ {\alpha_S}  \label{eqn:lossNS}
\end{equation}
when training at the critical batch size $B_{crit} = \frac{B_*}{L^{1 / \alpha_B}}$, and where $\alpha_N, \alpha_S, \alpha_B, N_c, S_c, B_*$ are all model-specific constants. 

Our goal is to minimize the total compute to train a model \new{of a given size $N_{target}$ } to a given target loss $L_{target}$.
We assume we have access to perfect growth operators that are loss-preserving and training dynamics-preserving, and that can grow from any model size to any other size\footnote{We can restrict the model size increases by adding additional constraints.}.  
We consider a training regime consisting of a number of stages.
In each stage $k$ we train a model with size $N_k$ for $S_k$ gradient steps, with the goal of \new{reaching size $N_{target}$ and} achieving a target loss $L_{target}$ at the end of the final stage.  We assume that $N_k \ge N_{k-1}$, and that there exists some way to initialize the model with size $N_k$ from one of size $N_{k-1}$ without changing the loss.  For simplicity, we neglect the batch size contribution to compute, and assume training is always at the critical batch size $B_* / L_{target} ^ {1/\alpha_B}$.

With these assumptions the total compute at the end of training for $M$ stages is:
\begin{equation}
    C = \sum_{k=1}^M 6 N_k \frac{B_*}{L_{target}^{1/\alpha_B}} S_k   \label{eqn:total_compute}
\end{equation}

We can compute the loss at the end of each stage in an iterative fashion.  The loss at the end of the first stage is given by $L_1(N_1, S_1)$ from Eqn.~\ref{eqn:lossNS}.  Then for each subsequent stage, we assume the loss curves can be translated 
and the loss at the end of stage $k$ is computed by starting with the loss at the end of the prior stage and decreased for $S_k$ steps.  To do so, first compute the effective number of steps $S_{eff,k}$ needed to reach initial loss for the stage $L_{k-1}$ with model size $N_k$, and then compute the loss at the end of the stage by $L_k(N_k, S_{eff,k} + S_k)$.
In summary:
\begin{eqnarray*}
  L_1 & = & \left( \frac{N_c}{N_1} \right) ^ {\alpha_N} + \left( \frac{S_c}{S_1} \right) ^ {\alpha_S} \\
  L_k & = & \left( \frac{N_c}{N_k} \right) ^ {\alpha_N} + \left( \frac{S_c}{S_{eff,k} + S_k} \right) ^ {\alpha_S}, k > 1 \\
  S_{eff,k} & = & \frac{S_c}{(L_{k-1} - (N_c / N_k) ^ {\alpha_N}) ^ {1 / \alpha_S}}.
\end{eqnarray*}

With this in hand, we can use a constrained optimizer to solve for the optimal schedule by minimizing Eqn. \ref{eqn:total_compute}, subject to the constraint that the \new{ final model size is the target size, and }loss at the end of training is the target loss. Formally,
\begin{equation*}
   \min_{\{(N_k, S_k)\}} \sum_{k=1}^M 6 N_k \frac{B_*}{L_{target}^{1/\alpha_B}} S_k
\end{equation*}
subject to
\begin{eqnarray*}
 L_M & = & L_{target} \\
 N_M & = & N_{target} \\
 0 & < & N_1 \\
N_{k-1} & \le & N_k, k > 1 \\
 0 & \le & S_k
\end{eqnarray*}

Note that in the single stage case ($M=1$)  \new{ with no $N_{target}$ condition},  this formulation reduces to the optimal calculation in~\appref{sec:optimal_schedule_appendix} of \citet{kaplan2020scaling}
\new{to find the optimal model size to reach a target loss. This} matches our training to \optimality. 

\new{
\paragraph{Measuring compute saving}

To measure the compute saving from staged training to reach a certain $L_{target}$, we use a single staged training ($M = 1$)
with no $N_{target}$ condition, and use our optimization algorithm to find the optimal compute and model size; this becomes $N_{target}$. 
Next, we run the optimization problem with $N_{target}$, $L_{target}$, and $M > 1$ to get the optimal training schedule and the expected 
compute, which we compare with  $M = 1$ to find the expected compute saving.

}


\begin{table}
\centering
    \small
  \def\arraystretch{1.12}\tabcolsep=6pt    
\begin{tabular}{ll}
                      \toprule
Number of stages               & Compute reduction factor  \\ \toprule
1 & 1.0 \\
2 & 0.83 \\
3 & 0.792 \\
4 & 0.779 \\
5 & 0.771 \\
10 & 0.763 \\

\bottomrule
\end{tabular}
\caption{Decrease in compute costs over an optimal single stage training regime.}
\vspace{-3mm}
\end{table}

\begin{table}
\centering
    \small
  \def\arraystretch{1.12}\tabcolsep=6pt    
\begin{tabular}{lcccc}
                      \toprule
Stage & $N_k$ & $S_k$ & $L_k$ & $C_k$ \\ \toprule
1 & 2.7M & 15.4K & 4.09 & 0.0033 \\
2 & 22.1M & 24.3K & 3.58 & 0.0154 \\
3 & 71.9M & 30.8K & 3.31 & 0.0365 \\
4 & 163M & 36.0K & 3.13 & 0.0665 \\
5 & 306M & 40.5K & 3.00 & 0.1056 \\

\bottomrule
\end{tabular}
\caption{Sample optimal schedule for a five stage regime to train to $L_{target}=3$ showing the number of non-embedding parameters $N_k$, the number of gradient steps $S_k$, the loss at the end of the stage $L_k$, and the compute in the stage $C_k$.}
\vspace{-3mm}
\end{table}

\paragraph{Observations}
An implementation of this optimization shows that the increase in compute efficiency for using multiple stages is independent of the target loss $L_{target}$, and quickly approaches $~0.76$ as the number of stages increases (using the scaling parameters for autoregressive transformer language models).

We also found that constraining the ratio between  consecutive  model sizes ($\frac{N_k}{N_{k-1}}$) to be 2, 4 or 8, leads to almost the same compute savings. These constraints come from the practical constraints of our implementation of the growth operators.


\new{

\section{Practical Stage Schedule}
\label{sec:practical_schedule_appendix}

The empirical values we estimated for the constants are: 

\begin{equation*}
    \begin{aligned}
    \tau_{opt} &= -0.052\\
    \tau_{depth} &= -0.0575\\
    \tau_{width} &= -0.0475\\
    \tau_{depth-width} &= -0.03\\
    \rho_{depth} &= 0.70\\
    \rho_{width} &= 0.55\\
    \rho_{depth-width} &= 0.40
    \end{aligned}
\end{equation*}
}
\end{document}